\documentclass[10pt,twocolumn,letterpaper]{article}

\usepackage[pagenumbers]{cvpr} 

\usepackage{graphicx}
\usepackage{amsmath}
\usepackage{amssymb}
\usepackage{booktabs}
\usepackage{cuted}
\usepackage{bbm}
\usepackage{scalerel}
\usepackage{xcolor}
\usepackage{amsmath,bm}
\usepackage{algorithm2e}

\DeclareMathOperator*{\argmax}{argmax} 
 
\usepackage[pagebackref,breaklinks,colorlinks]{hyperref}

\usepackage[capitalize]{cleveref}
\crefname{section}{Sec.}{Secs.}
\Crefname{section}{Section}{Sections}
\Crefname{table}{Table}{Tables}
\crefname{table}{Tab.}{Tabs.}

\begin{document}

\title{ Expressive Talking Head Video Encoding \\in StyleGAN2 Latent-Space}

\author{Trevine Oorloff and Yaser Yacoob\\
University of Maryland, College Park, MD 20742, USA\\
{\tt\small \{trevine,yaser\}@umd.edu}
}
\maketitle

\begin{strip}\centering
\includegraphics[width=0.95\linewidth]{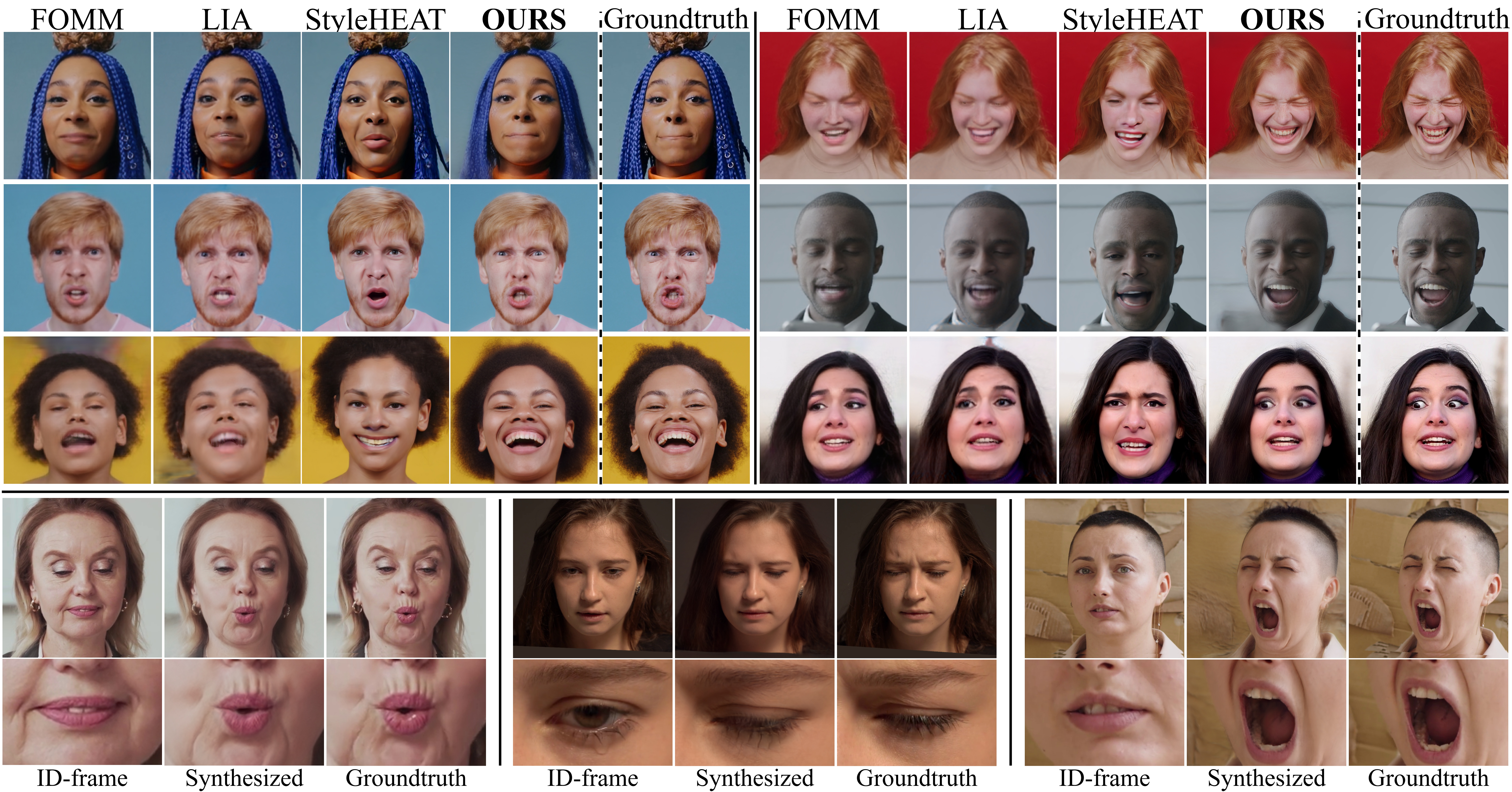}
\captionof{figure}{\label{fig:teaser} \textbf{The proposed framework is capable of capturing fine, detailed, and highly expressive facial features} (\eg, lip-pressing, mouth puckering and gaping, gaze, wrinkles). \textit{Top:} Demonstrates how our re-synthesis results compare with a few state-of-the-art models: FOMM \cite{siarohin2019first}, LIA \cite{wang2022latent}, and StyleHEAT\cite{yin2022styleheat}. \textit{Bottom:} Depicts the zoomed-in images of the ID-frame and the Synthesized frame generated through our approach using the encoding of the ID-frame (ID-latent) and 35 parameters capturing the facial deformations of Groundtruth.}
\end{strip}

\begin{abstract}

While the recent advances in research on video re-enactment have yielded promising results, the  approaches fall short in capturing the fine, detailed, and expressive facial features (\eg, lip-pressing, mouth puckering, mouth gaping, and wrinkles) which are crucial in generating realistic animated face videos. To this end, we propose an end-to-end expressive face video encoding approach that facilitates data-efficient high-quality video re-synthesis by optimizing low-dimensional edits of a single Identity-latent.
The approach builds on StyleGAN2 image inversion and multi-stage non-linear latent-space editing to  generate  videos that are nearly comparable to input videos.
While existing StyleGAN latent-based editing techniques focus on simply generating plausible edits of static images, we automate the latent-space editing to capture the fine expressive facial deformations in a sequence of frames using an encoding that resides in the Style-latent-space (StyleSpace) of StyleGAN2. The  encoding thus obtained could be super-imposed on a single Identity-latent to facilitate 
re-enactment of face videos at 1024\textsuperscript{2}. The proposed framework economically captures face identity,  head-pose, and complex expressive facial motions at fine levels, and thereby bypasses training, person modeling, dependence on landmarks/ keypoints, and low-resolution synthesis which tend to hamper most re-enactment approaches. 
The approach is designed with maximum data efficiency, where a single $W+$ latent and 35 parameters  per frame enable high-fidelity video rendering. This pipeline can also be used for puppeteering (\ie, motion transfer). A high-quality 4K-video dataset was used and will be released.  The project page is located at \url{https://trevineoorloff.github.io/ExpressiveFaceVideoEncoding.io/}.
\end{abstract}

\begin{figure*}[t!]
  \centering
  \includegraphics[width=0.72\linewidth]{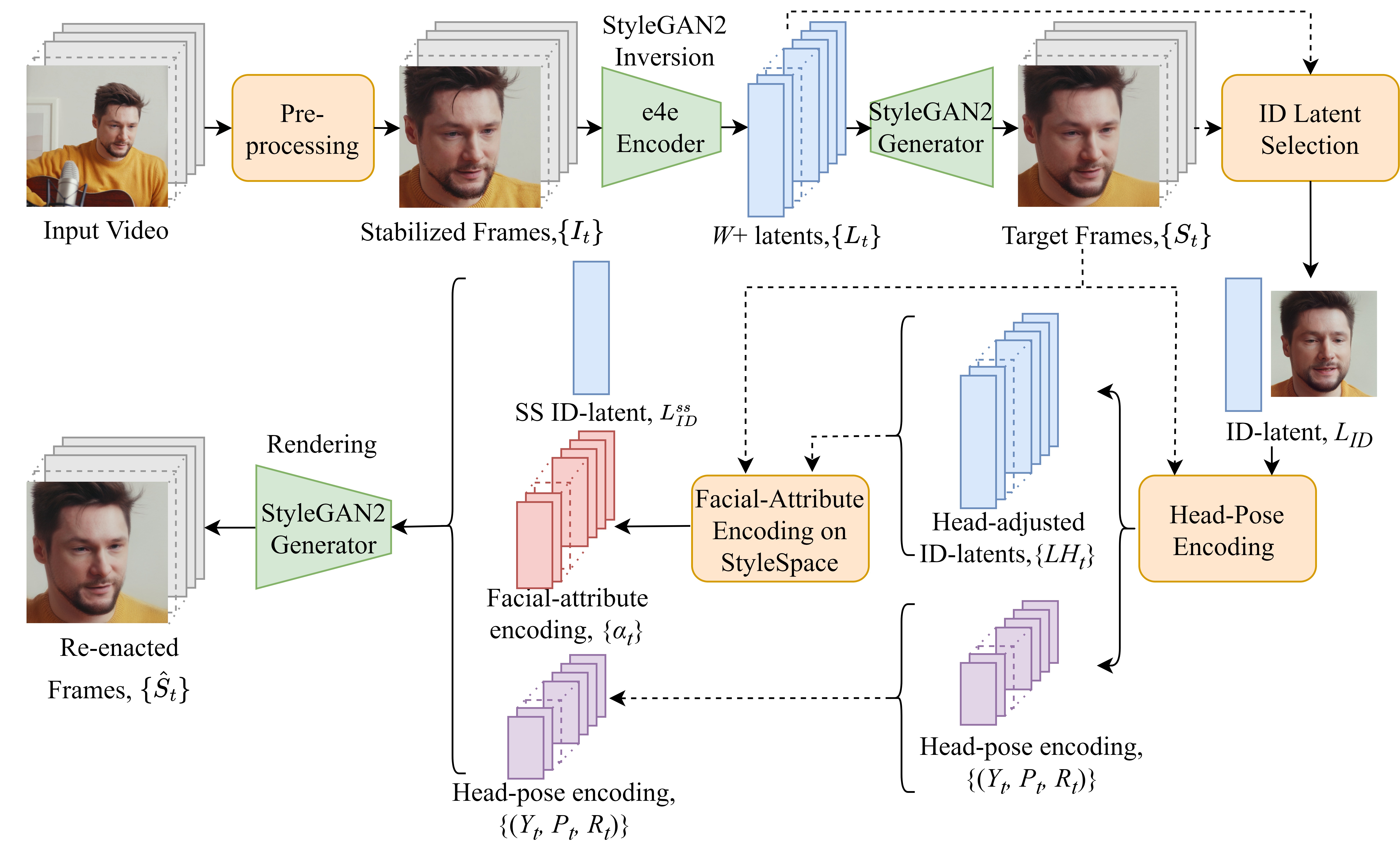}
  \caption{\label{fig:pipeline} \textbf{The multi-stage pipeline for encoding a video in latent-space.} 
  The \textit{(1) Pre-processing} stage aligns the input sequence of frames, which are fed to the \textit{(2) GAN inversion} stage to obtain the corresponding sequence of $W+$ latents. Out of which the best inversion which also has near frontal head-pose is chosen to be the ID-latent in the \textit{(3) ID-latent selection} stage. The \textit{(4) Head-pose encoding} stage, encodes the yaw and pitch of the target frames, in reference to the ID-latent while generating a series of head-pose adjusted ID-latents. Subsequently, the \textit{(5) Facial-attribute encoding} stage, encodes the facial deformations using 32 parameters anchoring onto the head-pose adjusted ID-latents. Finally, the encoded parameters (35/frame) and the ID-latent is used to synthesize the re-enacted frames at the \textit{(6) Rendering} stage. 
  }
  \vspace{-0.1in}
\end{figure*}

\section{Introduction}\label{sec:intro}

Talking-head re-enactment, which involves animating a static portrait image to mimic the changes in head-pose and other facial attribute deformations of a driving video while maximally preserving the identity across the frames, has a wide range of applications such as AR/VR, telepresence, and movie production. Intuitively to facilitate re-enactment, one has to decompose the motion from the identity of the driving sequence of frames, and to this end, most contemporary methods utilize facial landmarks/keypoints-based \cite{wang2021one,siarohin2019first,wang2019few}, 3D facial representation-based \cite{doukas2021headgan,zhang2021facial}, and latent-based \cite{wang2022latent} approaches to encode the facial deformations. While these methods generate promising results and each of them has its own pros and cons (\cref{sec:related}), most common drawbacks of the existing approaches include limitation to low resolution (commonly $256^2$ and $512^2$ at most), the requirement of extensive training data and person modeling, and especially the inability to capture extreme poses and intricate expressive facial details (see \cref{fig:teaser}) which detracts from the realism of re-enacted videos.   

On the other hand, the recent advances in StyleGAN2-based inversion techniques \cite{abdal2019image2stylegan,alaluf2021restyle,alaluf2021hyperstyle,roich2021pivotal,tzaban2022stitch} enable  manipulation of high-resolution ($1024^2$) real-world images \cite{abdal2020image2stylegan++,abdal2021styleflow,alaluf2021only,harkonen2020ganspace,shen2020interpreting,wu2021stylespace} due to the highly disentangled property of their latent-spaces. However, such latent-based manipulation techniques are mostly limited to static images and focus on simply generating plausible edits (\eg, changes to smile, age, hair color, \etc). While recent research \cite{bounareli2022finding,kang2022megafr,yin2022styleheat} has employed StyleGAN2 to generate high-resolution re-enactment video, they utilize 3D parametric models to capture facial deformations. While such priors are able to capture global facial attributes, they are not capable of capturing the fine and intensely expressive facial deformations.

In order to  bridge the gap between high-fidelity static portrait image synthesis/manipulation and face re-enactment of intense expressions and speech, we propose a novel end-to-end face video encoding approach that automates the latent-editing process to capture head-pose and fine and complex expressive facial deformations using merely 35 parameters/frame that reside in the Style-latent-space (StyleSpace, $SS$) of StyleGAN2.  We extend single image generation models, namely StyleGAN2 \cite{karras2020analyzing} and StyleFlow \cite{abdal2021styleflow} to the temporal dimension. Quantitative evaluation of latent-spaces: $Z,\,W,\,W+,$ and $S$, by \cite{wu2021stylespace}, indicates that within the StyleGAN2's latent-spaces, the proposed StyleSpace  has the best disentanglement, completeness, and informativeness. Thus, we perform edits on $SS$ as it enables control of individual facial attributes without re-training a network to enforce disentanglement \cite{donahue2017semantically}. Moreover, since the latent-spaces are sparse (\ie, only specific points in the space are visually valid and meaningful) we propose optimization frameworks that anchor the latent-space attribute editing to the real images. The computed latent paths between frames are non-linear and therefore avoid the limitations of common linear latent editors \cite{wang2022latent}. 

In this research, we focus on both re-synthesis and puppeteering of videos using a compact encoding scheme while focusing on accurate reconstruction of expressive facial deformations. In re-synthesis, we encode a video using a low-dimensional representation of small edits of an Identity-latent (ID-latent). The proposed pipeline is capable of capturing and regenerating complex facial features as shown in \cref{fig:teaser,fig:results} while achieving state-of-the-art performance at $1024^2$. Further, since the encoding is independent of the subject in the video, we can substitute the ID-latent (\eg, an inversion of a real face) of a different subject and apply the face deformation parameters to generate high-fidelity puppeteering video. 

Our video clip encoding is extremely compact, a single latent ($  18\times512$) corresponding to an ID-latent and only 35 parameters per frame that control the head-pose (3 parameters) and the facial features edits (32 parameters). The 70 bytes per frame are nearly half the state-of-the-art (see Tab. 5 \cite{wang2021one}). The key contributions of the paper are:
\begin{itemize}
     \setlength{\itemsep}{2.5pt}
    \setlength{\parskip}{0pt}
    \item a novel algorithm for high-resolution $(1024^2$) face video encoding 
    for re-synthesis and puppeteering with emphasis on precise reconstruction of both expressive and talking facial attributes in contrast to common models that only facilitate talking head re-enactment,  
    \item an extremely compact encoding  of head-pose and facial attribute deformations,
    \item a generative approach that employs optimization frameworks instead of person-centric data  training,
    \item a framework employing image inversion to anchor the sparse latent-edits to image-based constraints, facilitating accurate spatio-temporal modeling, 
    \item a novel method to find StyleSpace channels corresponding to facial attributes based on index sensitivity,
    \item an approach that automates the latent-space editing process to capture facial attribute deformations 
    in contrast to prevailing work on latent-space editing that simply illustrate plausible semantic visual results,
    \item a high-quality (4K) video dataset that will be shared.
\end{itemize}

\section{Related Work}\label{sec:related}
\subsection{Latent-Space Based  Editing}

Understanding the latent-space of a pre-trained GAN has led to better controllability over the generated output.
Research such as \cite{harkonen2020ganspace,shen2020interpreting} explore the latent-space of StyleGAN to identify the interpretable semantic directions that control attributes such as aging, smile, gender, pose, \etc within the latent-space. 
However, the entangled nature of the latent-space limits the manipulation, as it often leads to undesirable artifacts. 

StyleSpace \cite{wu2021stylespace}, StyleFlow \cite{abdal2021styleflow}, and StyleRig \cite{tewari2020stylerig} are a few prominent algorithms based on the StyleGAN2 architecture that yield  impressive control over latent-based manipulations. The authors of StyleSpace analyzed  $SS$ and formulated an algorithm to identify the style channels that control specific attributes by backtracking gradients. StyleFlow, on the other hand, uses a flow-based model conditioned on the attributes to enable non-linear and conditioned latent-space edits. Even though the StyleRig algorithm enables a rig-like control over the 3D semantic parameters of faces generated through StyleGAN, it has limited manipulative directions \cite{tewari2020stylerig}. In contrast to all these latent editing approaches, which simply generate plausible edits to static images, our algorithm attempts to automate the latent editing to quantify facial deformations in the form of $SS$ edits.

\subsection{Face Video Re-enactment}

Controlling the facial attributes and their motion through facial keypoints/landmarks are popularly used in video re-enactment \cite{gu2020flnet, wang2019few, zakharov2019few, zakharov2020fast, siarohin2019first}. While these approaches provide a strict guidance over the facial attributes, they are challenged to capture fine expressive facial details (\eg, hair, teeth, lip compression, wrinkle dynamics) and accessories (\eg, eyeglasses). Further, they are dependent on the accuracy of the landmarks and suffer in re-enactment video synthesis when the head and/or face geometries  of the source and target considerably differ \cite{tripathy2021single}.
Wang \etal in \cite{wang2021one} sought to address the latter through a 3D-keypoint representation which is in turn used to warp the source frame. Even though this method yields promising results, it requires extensive training (data and computational resources), the resolution is limited to $512^2$, and the generated frames lack sharpness and contain blocking artifacts.

Approaches such as \cite{doukas2021headgan,fried2019text,geng2018warp,nagano2018pagan,zhang2021flow} employ  3D facial models (\eg, 3DMM) to guide the synthesis, and are excellent at capturing facial movements. 
Despite the potential of 3D model-based approaches to generate high-quality videos, they represent only the inner-face region; thus are comparatively poor at constructing surrounding facial features (\eg, hair) or complex features such as teeth, wrinkles, complex mouth motion and require 3D training data that are resource and computation intensive. 

\subsection{StyleGAN-based Video Synthesis}

The ability to synthesize high-resolution photo-realistic images and the rich latent-space of StyleGAN are stimulating video synthesis research. \cite{tian2021good} and \cite{fox2021} train a temporal architecture that is used to navigate the latent-space of a pre-trained StyleGAN to search for temporally coherent directions for synthesizing videos at $1024^2$. While the former is limited to generating random video clips, StyleVideoGAN facilitates re-enactment using a PCA-based approach to transform the learned motion trajectories to the source image. While Bounareli \etal in \cite{bounareli2022finding} propose a method to find controllable directions of the $W+$ space of StyleGAN2 with the help of a 3D model synthesizing videos at $256^2$, the research of \cite{yin2022styleheat,kang2022megafr} utilize 3D models to capture the facial motion, hence share the drawbacks of 3D prior based models discussed above, despite the ability of generating $1024^2$ videos.

In addition to the inability of capturing the highly expressive facial attributes precisely, all these approaches attempt to learn a model that decomposes the motion-related content and hence require a training phase. 
In contrast,  our model extends the inherent disentangled nature of the StyleSpace ($SS$) of a pre-trained StyleGAN2 to achieve this decomposition in our pipeline. 
Further, in contrast to the above StyleGAN-based approaches which require  the entire latent ($18\times512$) per frame, the proposed framework provides an extremely compact encoding scheme comprising of 0.38\% of parameters per frame (35 vs.$18\times512$) while generating videos at $1024^2$.

\section{Methodology}

Our approach consists of six  stages: video pre-processing, GAN inversion, ID-latent selection, head-pose encoding, facial attribute encoding, 
and rendering. The entire flow is represented in \cref{fig:pipeline} and utilizes the e4e encoder \cite{tov2021designing}, StyleFlow, and StyleSpace
(with significant changes to these components to achieve our objectives). 

We use the following notation to describe the pipeline. 
Notations beginning with $L$ and $L^{ss}$ denote $W+$ latents and the corresponding $SS$ latents, respectively. 
$L^{ss}$ is obtained using the affine transform $\mathcal{A}( \cdot )$, \ie, $L^{ss} = \mathcal{A}( L )$. $I$ denotes a real image (groundtruth) and  $S$ denotes a synthesized image from a latent. For example, $S_t=G(L_t)$ describes the generation of an image from a latent, and the subscript refers to the frame at time $t$. $G$ is the original StyleGAN2 generator, but it is supplemented by two style generators, $G_{sf}$ for StyleFlow and $G_{ss}$ for StyleSpace.
$E$ is the e4e encoder used for real image inversion into $W+$ space. $Y_t$ and $P_t$ are the optimal Yaw and Pitch used by $G_{sf}$ at time $t$. Finally, $\mathbf{\alpha}_t$ is a 32-dimensional vector that controls the facial deformations of the generator $G_{ss}$, given a latent $L_t$.

The pre-processing stage  generates a set of face images that are stabilized and aligned so that their inversion to latent-space achieves maximal identity preservation and continuity of spatio-temporal head and face motions. The inversion employs the e4e encoder to generate a sequence of latents, $L_1,\dots,L_t$ in $W+$ space corresponding to the sequence of frames.
The images generated from these latents serve as the basis for rigid and non-rigid optimizations, replacing the raw image input. They enable controlled editability in conjunction with image loss metrics. It is important to note that optimization constraints are applied in the image space and not in the latent/parameter space. 
In the third stage, a single latent from the sequence is selected as an ID-latent, $L_{ID}$, for generating the various head-poses of the person in the video. 
\begin{equation}\label{EQ_ID}
    L_{ID}= \argmax_{L_t} (ID_{similarity}(I_t,G(L_t))
\end{equation}
Using a single $L_{ID}$ as the anchor to perform head-pose and facial motion edits, not only reduces the data requirement of rendering but also minimizes the identity variation across frames. In a re-enactment setting, the image corresponding to $L_{ID}$ functions as the single source image and the sequence of frames $\{I_t\}$ function as the driving frames. 
Refer \cref{sec:gan_inv,sec:id_latent} for further details on stages 2 and 3 respectively. 

The fourth stage finds, for each frame, the head transformation (\ie, $Y_t$ and $P_t$) in  StyleFlow latent-space  needed to render $L_{ID}$ as close as possible to the synthesized image $G(L_t)$
by minimizing,
\begin{equation}\label{EQ_SF}
     \min_{Y_t, P_t} \; \mathcal{L}\{G_{sf}(L_{ID} ,Y_t,P_t) \;, \;G(L_t)\}.
\end{equation}
$G_{sf}(L_{ID},Y_t,P_t)$ results in a new latent, ${LH}_t$, that captures the correct head-pose at time $t$ starting from $L_{ID}$.
The fifth stage solves, for each frame, the set of facial deformations $\alpha_t$  in $SS$, that when applied to ${LH}_t^{ss}$ matches as close as possible to $G(L_t)$. The result is a set of 32 parameters that achieve  $G(L_t)\approx G_{ss}(LH_t^{ss},\alpha_t)$ through minimizing,
\begin{equation}\label{EQ_SS}
    \min_{\alpha_t} \; \mathcal{L}\{ G_{ss}({LH}_t^{ss} ,\alpha_t) \;,\; G(L_t)\}.
\end{equation}

Finally, an image is synthesized by applying,
\begin{equation}\label{EQ_R}
S_t=G_{ss}(LH_{t}^{ss},{\alpha_t}) = G_{ss}(G_{sf}(L_{ID},Y_t,P_t),{\alpha_t}).
\end{equation}
Thus, synthesizing the re-enacted frame at time $t$ using a fixed $L_{ID}$ and 34 style controlling parameters (plus the initial Roll angle, $R_t$ used in pre-processing).

\subsection{Video Pre-Processing}

Face alignment is an important step in face image inversion regardless of whether an encoder or optimization approach is employed since StyleGAN2 is a fixed-resolution architecture. 
Temporal consistency of the alignment is critical due to the role each frame plays in our optimizations. Slight misalignments may change identity, head-pose, or misinterpret facial feature attributes (shape and dynamics). The alignment used in StyleGAN2 depends on the commonly used 68 facial landmarks, including mouth and eye coordinates for warping. However, these undergo dynamic changes in a video clip which generate jitters and rescaling in face alignment. To avoid the impact of dynamic coordinates, \cite{fox2021} cropped the full face excluding the eyes and mouth coordinates.
We consider this insufficient to alleviate the combined effects of head-pose and facial motions. Instead, our alignment aims to: (1) completely stabilize the head when head-pose does not change between consecutive frames, so that non-rigid face motions are captured in a maximally aligned form, (2) rely on inversion to capture the relative head alignment when the head-pose rotates out-of-plane. 

We employ \cite{openface2018} for detecting faces and tracking features in a video clip.  However, the landmarks are not sufficiently accurate for face alignment over a sequence of frames. Dense optical-flow captures a combination of rigid and non-rigid facial motions. However, since our objective is to only align the rigid head motion between frames, we employ a parametric optical-flow model \cite{black1997} to register a frame at time $t$ to a key frame $k_i$ at time $i$ ($<t$). When the rigid head motion is small or limited to the 2D plane, the registration is accurate for the duration (occasionally, several tens of frames), but upon out-of-plane head rotation, the registration requires adjusting the key frame to a new $k_{i+1}$. When the head out-of-plane rotation is rapid, consecutive frames may become key frames. 
(Further details in \cref{supp_sec:alignment}).

\subsection{Head-Pose Encoding}

Temporally consistent head-pose is challenging to recover and synthesize. Head-pose is represented by three degrees of rotation, Yaw, Pitch, and Roll, computed with respect to a virtual point at the center of the head. 
There are numerous landmark and mesh-based approaches for estimating head-pose. Instead, we choose an analysis-by-synthesis approach to estimate the closest rendering of a latent to the target image (\cref{EQ_SF}). StyleFlow  proposed an effective system for a single latent-based edit of head-pose by controlling the Yaw and Pitch angles. The Roll angle is a 2D image-based transformation and is relegated to a pre-processing step necessary for face-alignment as required by StyleGAN2. 

An important feature of StyleFlow is that the attribute editing direction is dependent and conditioned on the given latent (\ie, it is specific to a person and relevant attributes captured by the generator).  This conditional architecture leads to improved disentangling and it also allows continuous parameter editing. Critically, the edit path is non-linear in the latent-space in contrast to the state-of-the-art that relies on linear and fixed directions that apply to all latents. 
 
We re-formulate the head-motion as a head-pose matching problem between a rendered image of the real-frame's encoded latent, $L_t$, and the rendered image of a rotated $L_{ID}$ which is solved as a minimization problem (\cref{EQ_SF}). The minimization employs two losses,  L2  and  LPIPS  to search the Yaw-Pitch space using gradient descent. These losses are computed over a masked area of the face that is based on an 81-landmark model (an extension of the 68-landmark model to include the forehead). However,  the eyes, mouth, and eyebrows are excluded in the  L2  loss, since these non-rigid areas are not relevant to 3D head rotations. 
The outcome of this stage is an alignment of the $L_{ID}$ to match the head-pose at time $t$, and it is represented by a new latent $LH_t$ (in $W+$) that will be further edited to capture the non-rigid motions of the eyebrows, eyes, mouth, and chin.

\subsection{Facial Attribute Encoding}\label{sub:fac_attr}

The facial attribute encoding extends \cite{wu2021stylespace}, where the authors demonstrate the highly disentangled nature of the $SS$ and provide a few StyleSpace indices that have mostly disentangled control over facial attributes. 
The facial-attribute encoding, $\alpha_t$,  (32 parameters) of each frame is applied to the latent $LH^{ss}_t$, which is a transformation of $LH_t$ to $SS$  via $LH^{ss}_t = \mathcal{A}( LH_t )$. 

\subsubsection{Choice of StyleSpace Indices}

The StyleSpace indices are analyzed to  make sure that maximally disentangled indices that capture complex and detailed expressive facial attributes as shown in \cref{fig:teaser,fig:results} are selected. 
For a specific facial feature $f \in \mathcal{F}$, we score each index $i \in \{l,c\}$ using index sensitivity, $\Gamma_{f,i}$, which  measures the change in image space for a unit change in the StyleSpace index.
$\Gamma_{f,i}$ is defined as,
\begin{equation}\label{eqn:idx_sens}
    \Gamma_{f,i} = \frac{1}{|\{\delta_k\}|}\sum_k{ \left\{\frac{ \mathcal{L}_{LPIPS}(S_k*M, S_{k-1}*M)}{ | \delta_k - \delta_{k-1} |} \right\}},
\end{equation}
where $S_k = G_{ss}(L^{ss}_{ID} + \delta_k\mathbbm{1}_i)$ is the synthesized image generated using $L_{ID}$ perturbed by $\delta_k$ at $SS$ index $i$, $M$ is the binary mask over the facial attribute considered, and $\mathbbm{1}_i = $\{$1$  when $(l,c) = i$;  $0$ elsewhere\}.  We choose $\{\delta_k\}$ to be a sequence of successive values with $|\{\delta_k\}|$ elements, 
and the subscript $k$ indicates the iterating index. Additionally, we calculate the index sensitivity over the whole face (\ie, $M$ is a matrix of ones that covers the whole face) and is denoted by $\Gamma_i$.
Subsequently, we rank the indices based on $\Gamma_{f,i}$ and $\Gamma_i$ values and choose the indices that have a higher $\Gamma_{f,i}$ and a negligible $\Gamma_i$ based on simple thresholding. We repeat the scoring on multiple subjects and frames sampled from the dataset and obtain the prominent indices across the sampled data. This novel approach enables the selection of maximally disentangled StyleSpace indices corresponding to the specific facial attribute chosen. The list of facial attributes $\mathcal{F}$ and the set StyleSpace indices $\mathcal{V}$, thus chosen are tabulated in \cref{tab:SS_idx} in the Appendix. 

The significance of our $SS$ indices selection process as opposed to the algorithm proposed in \cite{wu2021stylespace} is as follows. We observed that the StyleSpace, $SS$ representation is not unique. \ie, optimizing 
\begin{equation}
    \min_{\alpha_{inv_t}} \; \mathcal{L}\{ G_{ss}({LH}_t^{ss}+\alpha_t+\alpha_{inv_t}), G_{ss}({LH}_t^{ss})\}
\end{equation}
does not necessarily yield $\alpha_t+\alpha_{inv_t}\approx0$.  
Therefore, as \cite{wu2021stylespace} back propagates to compute the gradient w.r.t. a $SS$ index, the gradients would be less accurate, as the $SS$ indices contributing to an identical facial deformation of two frames would differ (as not unique). Instead, we use a forward approach, perturbing each index separately and computing the corresponding deformation loss, thus directly computing the true gradient (sensitivity in the image space for a unit change of each $SS$ index) which is more accurate. 
 
\textbf{Facial Deformation Attribute Encoding:}\label{sec:FaceDeform}
We compute the optimal encoded latent values, $\alpha_t$, that edit facial attributes to capture the facial deformations. $\alpha_t$ represents the offset values from $LH^{ss}_t$ and is obtained through a per-frame optimization (\cref{EQ_SS}) over the $SS$ indices and is presented in \cref{algo:att_enc} in the Appendix. The reconstruction  of the latent $L_t$ obtained from the e4e encoder is  used as the groundtruth in the optimization and denoted by $S_t$, while the rendered re-enacted frame during the optimization is denoted by $\hat{S}_t$.  

\textbf{Initialization of indices ($\boldsymbol{LH^{ss}_t}$):} Due to the sparsity of the latent-space and as the optimization is over a multi-dimensional space, it is highly probable for the optimization algorithm to converge consecutive frames, which are nearby in image-space, onto local-minima that are distant in the latent-space. 
The slight differences in the optimum point of consecutive frames could introduce jitter in re-enactment. 
Therefore, to bias the algorithm to solve for $\alpha_t$ in the vicinity of the previous frame's optimum, we initialize the $SS$ indices we optimize, $i=(l,c) \in \mathcal{V}$ of $LH^{ss}_t$ as, 
\begin{equation}
    LH^{ss}_t(l,c) = LH^{ss}_{t-1}(l,c),\, \forall (l,c) \in \mathcal{V}.
\end{equation}
\textbf{Index-specific learning rate, $\boldsymbol{\eta_{f,i}}$:} We observed that different subjects and indices have different sensitivities to a unit change in the StyleSpace ($\Gamma_{f,i}$) (see \cref{sec:idx_sens}). This observation corroborates the non-linear nature of latent editing discussed in StyleFlow. Hence, using the same learning rate across all indices would result in an undue dominance of high-sensitivity indices, thus generating non-optimal results. Therefore, for each input video and each facial attribute, we compute an index-specific learning rate using,
\begin{equation}
    \eta_{f,i} = \exp\left\{-1.5 \,\Gamma_{f,i}\, / \max\limits_{i \in \mathcal{V}_f } \,(\Gamma_{f,i})\right\},
\end{equation} 
that was obtained empirically. For each epoch, optimization is done in parallel for all the attributes and the optimization over indices corresponding to the gaze is skipped for frames where blinking is detected.

\textbf{Loss Functions:} The algorithm is optimized by minimizing over multiple losses. The total loss is defined as,
\begin{align}
    \mathcal{L} = \mathcal{L}_m + \mathcal{L}_e + \mathcal{L}_p + \mathcal{L}_{ID} + \mathcal{L}_{FP} \label{eq:total_loss},
\end{align}
where the loss terms $\mathcal{L}_{ID}$ and $\mathcal{L}_{FP}$ represent the identity loss and the Face-Parsing loss respectively and the subscripts $m$, $e$, and $p$ correspond to the losses computed over extracted regions of the \{mouth + chin/ jaw\}, \{eyes + eyebrows\}, and \{pupil\}, respectively.
\begin{align}
      \mathcal{L}_m &= \mathcal{L}_{LPIPS_m} + \mathcal{L}_{L2_m} + \mathcal{L}_{IF_m}, \label{eq:loss_m}\\
      \mathcal{L}_e &= \mathcal{L}_{LPIPS_e} + \mathcal{L}_{L2_e} + \mathcal{L}_{IF_e}, \label{eq:loss_e}\\
      \mathcal{L}_p &= \mathcal{L}_{L2_p} + \mathcal{L}_{IF\_L2_p}, \label{eq:loss_p}
\end{align}
where $\mathcal{L}_{LPIPS}$, $\mathcal{L}_{L2}$, and $\mathcal{L}_{IF}$ represent the LPIPS loss, L2 loss, and Inter-frame loss, respectively. Refer \cref{sec:opt_details} for further details. 

\subsection{Rendering}
Once the encoding is complete, the $L_{ID}$ and the time-series of the 35 parameters, $\{\alpha_t, Y_t, P_t, R_t\}$   are transmitted to the renderer. To synthesize the re-enactment video, first $LH_t$ is obtained from $L_{ID}$ to adjust for the head-pose using StyleFlow for each frame. Then $LH_t$ is transformed to $SS$, and the 32 indices responsible for the facial attributes, $\alpha_t$ are applied to synthesize the image using the generator, $G_{ss}$.
\begin{equation}{\label{eq:ren}}
    \hat{S}_t = G_{ss}(LH^{ss}_t + \alpha_t\mathbbm{1})
\end{equation}

There exist an inherent quality loss in the initial encoding as the real-world subjects would mostly be out-of-domain of StyleGAN resulting in notable deviations between the e4e encoded frames and real frames. \cite{roich2021pivotal} propose fine-tuning the StyleGAN's generator around ``pivots" to improve the photo-realism of images while maintaining editability. Adapting from this concept, we fine-tune only the layers post-StyleSpace, $G_{ss}$ by solving \cref{EQ_Gft} using \{$LH^{ss}_t, \alpha_t$\} as pivots (in contrast to $W+$ latents in PTI) with real frames as reference.
The optimization is performed over the entire sequence of frames simultaneously compared to the single image tuning in PTI. 
\begin{equation}\label{EQ_Gft}
    \min_{\theta} \; \sum_t\mathcal{L} \{G_{ss}({LH}_t^{ss},\alpha_t; \theta) \; , \; I_t\} 
\end{equation}

\section{Experiments and Results}

\begin{table*}[ht!]
    \centering
    \resizebox{0.88\linewidth}{!}{
    \begin{tabular}{ l |  c c c c c c c c c c c } 
        \hline
        
        Method & res. & L1 $\downarrow$ & LPIPS$\downarrow$ & $\mathcal{L}_{ID}$$\downarrow$ & PSNR$\uparrow$  & SSIM$\uparrow$ & FID$\downarrow$ & FVD$\downarrow$  & $\rho_{\scaleto{AU}{3pt}}$$\uparrow$ &
        $\rho_{\scaleto{GZ}{3pt}}$$\uparrow$ &
        $\rho_{\scaleto{pose}{3pt}}$$\uparrow$\\  
        
        \hline
         
        Wang \etal * & $512^2$ & \underline{2.42} & \underline{0.030} & \underline{0.087} & \underline{32.8} & \underline{0.957} & \textbf{12.0} & \underline{82.3}  & \underline{0.881} & \underline{0.965} & \underline{0.983}\\
        
        StyleVid.GAN * & \textbf{1024\textsuperscript{2}} &4.04 & 0.109 & 0.104 & 28.8 & 0.926 & 28.8 & 223.3  & 0.739 & 0.884 & 0.979\\
        
        \textbf{Ours*} & \textbf{1024\textsuperscript{2}} & \textbf{1.96} & \textbf{0.026} & \textbf{0.067} & \textbf{34.1} & \textbf{0.960} & \underline{13.6} & \textbf{79.8}  & \textbf{0.899} &  \textbf{0.971}  & \textbf{0.987}\\
        
        \hline 
        
        FOMM & $256^2$ & \underline{3.07} & \underline{0.036} & 0.174 & \underline{31.0} & 0.932  & 28.7 & \underline{140.3}  & \underline{0.710} & 0.755 & 0.648\\  
        
        LIA  & $256^2$ & 3.24 & 0.042 & 0.164 & 30.0 & 0.929  & 30.2 & 162.9 & 0.546 & 0.693 & 0.619\\  
        
        fs-vid2vid  & $512^2$ & 5.75 & 0.093 & 0.158 & 25.2 & 0.900 & 42.4 & 359.6  & 0.571 & \underline{0.784} & 0.629 \\  
        
        StyleHEAT & \textbf{1024\textsuperscript{2}} & 4.13 & 0.097 & \underline{0.134} & 27.6 & \underline{0.933} & \underline{25.1} & 281.9  & 0.673 & 0.701 & \underline{0.763} \\
        \textbf{Ours} & \textbf{1024\textsuperscript{2}} & \textbf{1.99} & \textbf{0.030} & \textbf{0.097} & \textbf{34.2} & \textbf{0.963} & \textbf{15.9} & \textbf{85.2}  & \textbf{0.771} & \textbf{0.834} & \textbf{0.880} \\
        
        \hline 
         
        \textbf{Ours} (ReStyle) & $1024^2$ & 2.01 & 0.031 & 0.099 & 34.0 & 0.959 & 16.9 & 93.9  & 0.767 &  0.831  & 0.843 \\
        
        \textbf{Ours} -- PTI\textsubscript{post} & $1024^2$ & 2.71 & 0.048 & 0.127 & 32.0 & 0.956 & 23.2 & 125.7  & 0.726 & 0.819 & 0.833\\
        
        \hline
         
    \end{tabular}
    
    }
    \caption{\textbf{Quantitative comparison of video re-synthesis against baselines.} The \textit{top (*)}, \textit{middle}, and \textit{bottom} sections respectively consist of metrics computed for the 6 videos received upon requests to authors, evaluation against the dataset of 150 videos, and results of ablations. 
    We yield state-of-the-art results at $1024^2$ on all metrics while using only 0.38\% of latent-space parameters of StyleGAN2. } 
    \label{tab:re-synthesis}
    \vspace{-0.1in}
\end{table*}

\subsection{Dataset and Evaluation}\label{sub:dataset_eval}

We selected 150 video clips (4K videos) from the video-sharing site \url{www.pexels.com} that combine high visual quality with expressive head and facial motions that are present in common low-resolution datasets. 
Each video  contains a single face performing significant face deformations, head motion, and speech. 
Additional details on the dataset are in \cref{sec:dataset}. 

\begin{table}[t]
    \centering
    \resizebox{\linewidth}{!}{
    \begin{tabular}{ l  | c c c c c c }
        \hline 
        
        Method& res. &$\mathcal{L}_{ID}$$\downarrow$&FID$\downarrow$& FVD$\downarrow$ & $\text{FVD}_{\scaleto{M}{3pt}}$$\downarrow$ & $\rho_{\scaleto{AU+GZ}{3pt}}$$\uparrow$ \\
        
        \hline 
        
        FOMM  & $256^2$ & \underline{0.153} & 77.0 & \underline{396.8} & \underline{103.0} & 0.501 \\
        
        LIA  & $256^2$ & 0.174 & 82.3 & 406.0 & 112.4 & 0.527 \\
        fs-vid2vid  & $512^2$ & 0.202 & \underline{73.6} & 445.1 & 112.7 & 0.640  \\
        
        StyleHEAT  & \textbf{1024\textsuperscript{2}} & 0.181 & 81.0 & 437.5 & 109.8 & \underline{0.667}  \\
        
        \textbf{Ours} & \textbf{1024\textsuperscript{2}} & \textbf{0.095} & \textbf{63.9} & \textbf{386.5} & \textbf{82.3}  & \textbf{0.708} \\
        
        \hline
    \end{tabular}
    }
    \caption{\textbf{Quantitative comparison of puppeteering against baselines} evaluated across 50 puppet-puppeteer pairs. Our approach achieves the best performance across all metrics.}
    \label{tab:puppeteering}
    \vspace{-0.1in}
\end{table}

We compare our results against two state-of-the-art StyleGAN-based models (most relevant): StyleHEAT and StyleVideoGAN, a latent-based model: LIA, and three other state-of-the-art models (keypoint/landmark-based) that facilitate re-enactment: fs-vid2vid, FOMM, and Wang \etal \cite{wang2021one}. Publicly available models were used for all algorithms except StyleVideoGAN and \cite{wang2021one} for which the authors kindly processed six videos. Note: All algorithms were evaluated at their native resolution using multiple metrics scoring: spatial quality, spatio-temporal quality and appearance, and temporal consistency of identity (further details in \cref{sec:metrics}).

\begin{figure*}[tb]
  \centering
  \includegraphics[width=\linewidth]{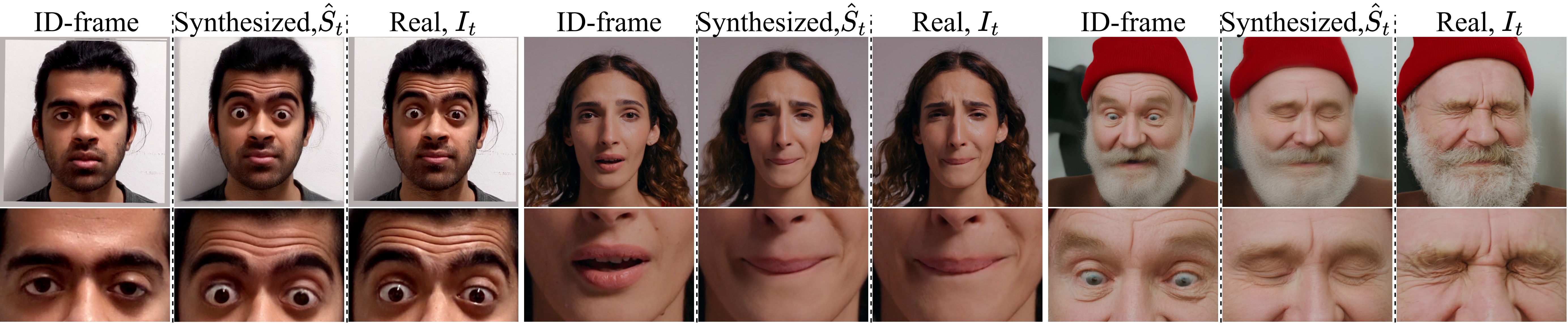}
  \caption{\label{fig:results} \textbf{Qualitative examples yielded through our approach} (in addition to \cref{fig:teaser}).  The StyleSpace indices and the optimization procedure were carefully designed such that complex and fine facial details such as lip-pressing, mouth puckering, mouth gaping, and wrinkles around the eyes, mouth, nasal-bridge, and forehead are well-captured. }
  \vspace{-0.1in}
\end{figure*}

\begin{figure}[t!]
    \centering
    \includegraphics[width=0.98\linewidth]{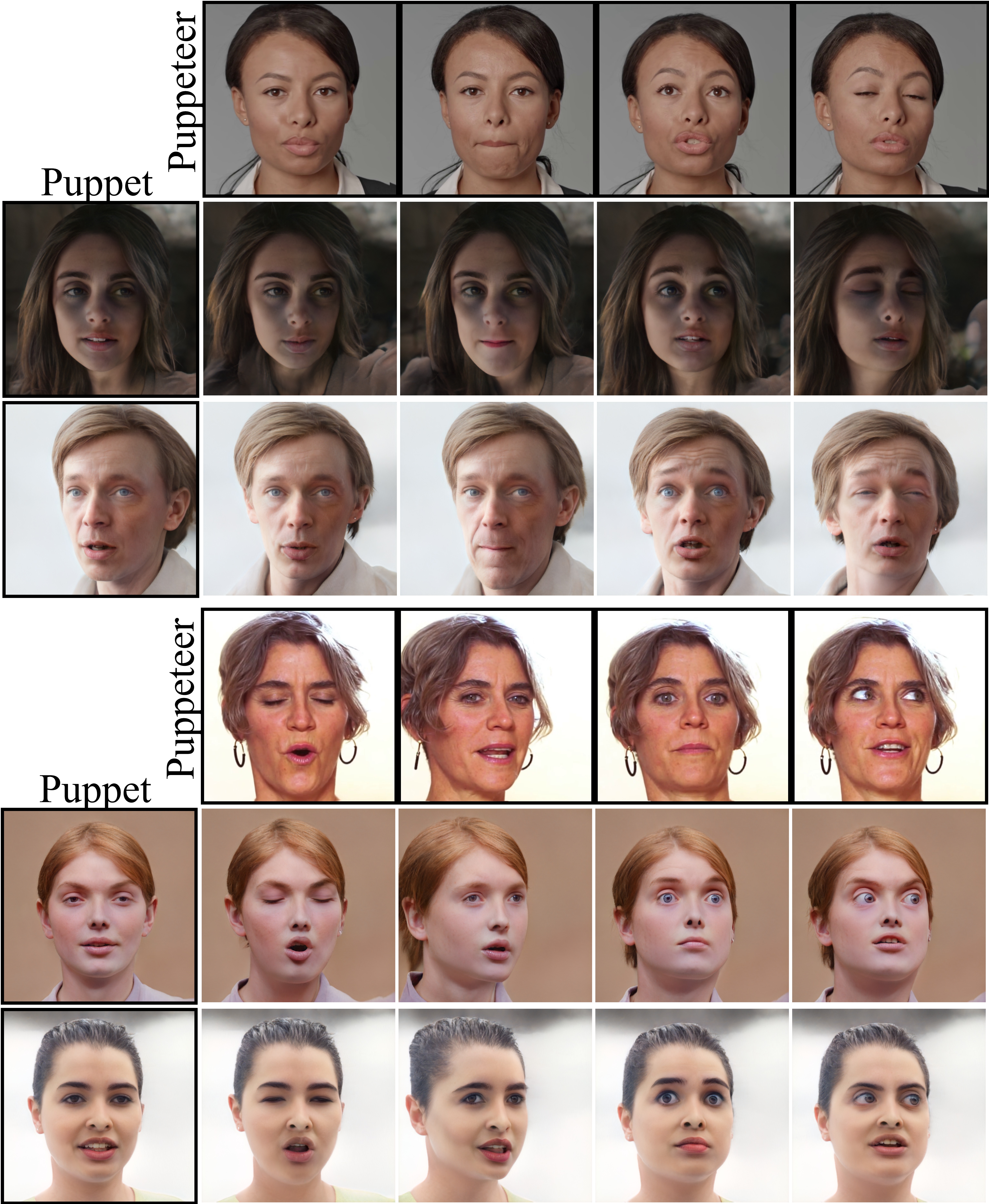}
    \caption{ \textbf{Qualitative evaluation of puppeteering}, where the encoded parameters of the puppeteer are applied to the ID-latent of the puppet. It could be observed that even complex facial deformations are transferred well across different identities.}
    \label{fig:puppeteering }
    \vspace{-0.1in}
\end{figure}

Referring to the top of \cref{tab:re-synthesis}, we achieve state-of-the-art performance at $1024^2$ with significantly improved re-synthesis results compared to StyleVideoGAN while utilizing only 0.38\% of the latent-space parameters used by them ($35$ vs. $18\times512$ per frame). Further, our encoding scheme outperforms Wang \etal in almost all metrics except FID (yet comparable) despite metrics of (Wang \etal) being computed in its native resolution of $512^2$. Moreover, evaluation across the full dataset, (middle of \cref{tab:re-synthesis}), shows that our approach outperforms fs-vid2vid, FOMM, LIA, and StyleHEAT in all scores by large margins. It is critical to note that lower native resolutions \cite{siarohin2019first,wang2019few,wang2021one,wang2022latent} significantly favor several metrics since there is no penalty for loss of details (\eg, L1, SSIM, FVD, \etc) with respect to $1024^2$ metrics.
Hence it is essential to emphasize on the qualitative analysis which more accurately reflects the potential of our framework. 

\cref{fig:teaser,fig:results}  illustrate,  qualitatively,  the  capturing of fine facial details such as lip pressing, mouth puckering and gaping, dynamic wrinkles around the eyes, mouth, nasal-bridge, and forehead enhancing photo-realism of the re-enacted videos which are not necessarily captured by the metrics 
(see \cref{fig:supp_resyn,fig:supp_p2p} and videos in the project page for more examples). 
To the best of our knowledge, such fine expressive details were not explicitly addressed by previous research. 

\begin{table}[t]
    \centering
    \resizebox{0.85\columnwidth}{!}{
    \begin{tabular}{  l | c  c  c  c c  } 
        \hline
        
        Method & Vid.1 & Vid.2 & Vid.3 & Vid.4 & Vid.5 \\ 
        
        \hline
        
        StyleFlow & 46.7  & 41.3 & 33.7 & 17.0 & 38.5 \\ 
        
        \textbf{Ours} & \textbf{16.0}  &  \textbf{19.9} & \textbf{16.3} & \textbf{11.5} & \textbf{21.9} \\
        
        \hline
    \end{tabular}
    }
    \caption{Mean head-pose loss ($\downarrow$) comparison between straightforward head-pose adjustment using StyleFlow vs. our approach on few sample videos. 
    }
    \label{table:SF_comp}
    \vspace{-0.1in}
\end{table}

Similarly, as shown in \cref{tab:puppeteering}, we achieve the best puppeteering results across all metrics considered. Further, \cref{fig:puppeteering } demonstrates the versatility of our method as even complex facial attribute deformations (\eg, lip pressing, puckering, wrinkles) of the driving frames are transferred successfully through the proposed framework. 

\subsection{Ablation Study}

As ablations, we study several design choices in our pipeline, namely: the use of a different encoder, the significance of the head-pose encoding approach, using real frames as reference in facial attribute optimization,  and the effect of the generator fine-tuning stage. 

Using ReStyle encoder\cite{alaluf2021restyle} replacing e4e generates comparable results (\cref{tab:re-synthesis}) implying that the proposed scheme is functional irrespective of the encoder provided that the inversion is within the editable domain of the latent-space.

Further using real frames $\{I_t\}$ as reference for the facial attribute encoding optimization (\cref{sec:FaceDeform}) instead of the synthesized frames $\{S_t\}$ resulted in visually sub-optimal results requiring us to abandon tighter pixel-level metrics as $\mathcal{L}_{L2}$, which are essential in capturing fine facial details such as wrinkles and gaze. Hence, we opted to use  $\{S_t\}$ for the optimization stage. We suspect this behavior to be caused due to the natural noise present in real images to which the StyleSpace optimization might be sensitive to. 

Even though StyleFlow is capable of directly generating a head-pose adjusted latent, provided \{$Y_t, P_t$\}, the quantified estimates of head-pose (using OpenFace) for a video stream are not sufficiently accurate to render using StyleFlow, resulting in inaccurate poses and significant jitter. Our synthesis-based optimization approach based on losses in image-space generates more accurate head-pose images consistent with reference frames (see \cref{table:SF_comp}).

It could be observed that while the re-synthesis results without the fine-tuning stage (\textbf{Ours}-PTI\textsubscript{post} in \cref{tab:re-synthesis}) yet outperform fs-vid2vid, FOMM, LIA, StyleVideoGAN, and StyleHEAT in almost all scores, the fine-tuning stage improves the photo-realism which is well reflected in the performance improvement with the addition of the fine-tuning stage. This is justifiable due to the tendency of real-world subjects being out of the domain of StyleGAN and the inherent loss of the encoder.

\subsection{Limitations}

Despite the promising results, the proposed approach has a few limitations. As the pipeline is based on the StyleGAN2 architecture, it inherits the limitations from StyleGAN2 and its inversion methods (\eg, fixed resolution, alignment requirements, texture sticking). 
While StyleGAN3 could alleviate the issue of texture sticking, we choose the richer latent space of StyleGAN2 due to its better disentangled latent space which is well structured and expressive for editing as opposed to
StyleGAN3 \cite{alaluf2022third}.
Further, the encoding pipeline is sensitive to occlusions resulting in visual artifacts in the synthesized images. Additionally, certain scenarios with extreme facial deformations and profile views could yet be challenging, which stems from the low representation of the FFHQ dataset used in training StyleGAN2.

\section{Conclusion}

We extend the StyleGAN2's photo-realism and disentanglement of its StyleSpace spatio-temporally, to propose a novel end-to-end pipeline for latent-based expressive face video encoding, which enables high-fidelity ($1024^2$) video re-enactment using a single $W+$ latent and 35 parameters per frame. 
Our algorithm achieves state-of-the-art performance while using a fraction (0.38\%) of parameters compared to StyleGAN latent-based approaches. 
To the best of our knowledge we are the first to (1) automate latent-space editing (that was previously used to merely generate plausible facial edits) to capture extremely fine, rich, and complex facial deformations, and (2) to propose an extremely compact latent-based face video encoding scheme based on StyleGAN2 enabling re-enactment. The negative societal impact is discussed in \cref{sec:negative}.

{\small
\bibliographystyle{ieee_fullname}
\bibliography{egbib}
}


\pagebreak
\appendix
\section*{Appendix}

\section{Overview}
\label{sec:overview}

The outline of the appendix is as follows.
\begin{itemize}
    \setlength{\parskip}{0pt}
    \item Sec.~\ref{supp_sec:alignment}: Detailed steps on alignment in the pre-processing stage
    \item Sec.~\ref{sec:gan_inv}: Discussion on the GAN inversion stage
    \item Sec.~\ref{sec:id_latent}: Additional details on identity-latent selection
    \item Sec.~\ref{sec:stylespace}: Illustrated explanations of noteworthy sections of facial attribute encoding
    \item Sec.~\ref{sec:results}: Further details and examples of experiments, results, and limitations of the proposed framework
    \item Sec.~\ref{sec:negative}: Discussion on Potential Negative Societal Impact
\end{itemize}

\begin{figure*}[h!]
  \centering
  \includegraphics[width=\linewidth]{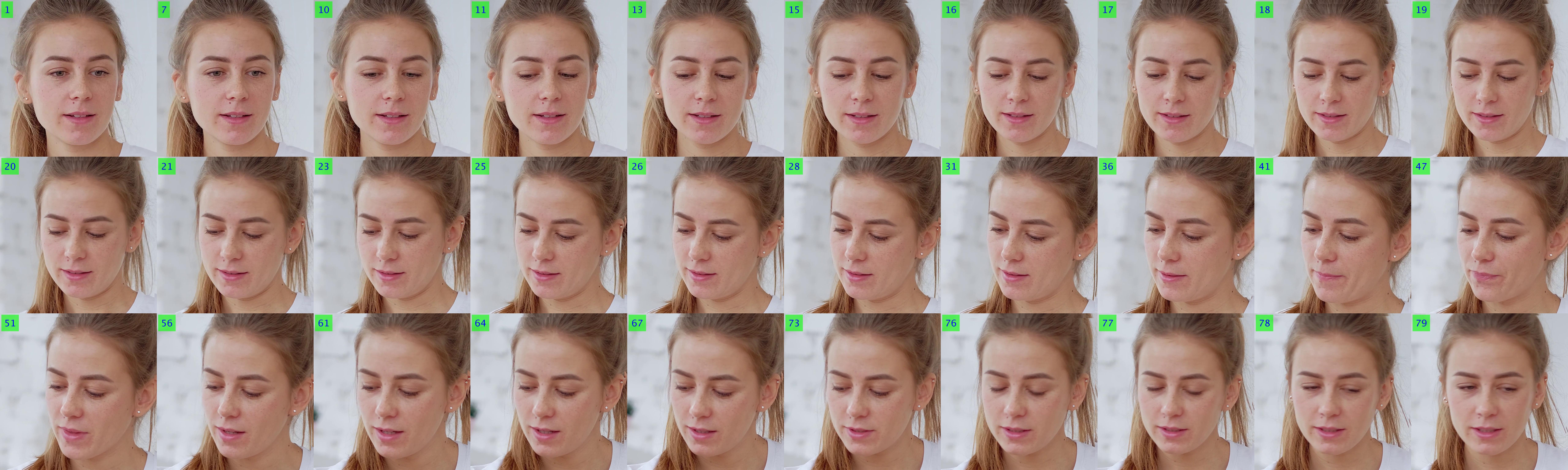}
  \caption{\label{supp_fig:PreProcess} \textbf{Key frames in a video sequence with head out-of-plane rotation}, where the head moves to near profile and then back. }
  \vspace{-0.1in}
\end{figure*}

\section{Video Pre-Processing: Alignment} \label{supp_sec:alignment}

The alignment carried out in the pre-processing stage could be elaborated further using the three steps below.
\begin{enumerate}
    \setlength{\itemsep}{2.5pt}
    \setlength{\parskip}{0pt}
    \item Detect eye blinking and compensate  for its effect on landmark location of the eyes. This improves StyleGAN2-based alignment by removing the sensitivity to eye shape change during blinking.
    \item Registration of the face between a frame and a key frame uses a parameterized affine optical-flow model of the head \cite{black1997}, excluding the non-rigid face features (eyebrows, eyes, and mouth). The over-constrained optical-flow model is very effective at stabilizing the face between consecutive frames unless there are changes in the Yaw/Pitch of the head. We employ the mean L2 distance to automatically determine the quality of the inter-frame alignment over the non-rigid parts of the face (\ie, compute the residual error in RGB values of face stabilization). A mean distance beyond a fixed threshold indicates that the affine motion model is not successful at stabilizing the rigid part of the face, triggering step (3).
    \item Key frame change that forces a new key frame to be the basis for future frames' face stabilization (aligned according to step (1)).
\end{enumerate}

For optical flow head registration, the threshold of the mean RGB registration error over the face (excluding eyes, mouth, and eyebrow areas) had to exceed 45.0 (if the inter-frame Yaw and Pitch change is less than $2^\circ$), or 30.0 (if the inter-frame Yaw or Pitch change exceeds $2^\circ$). The objective is to avoid forcing face registration when the head is moving out-of-plane. Instead, a change in the key frame is triggered, allowing the StyleGAN2 encoder to capture the new head-pose.
\cref{supp_fig:PreProcess} depicts the key frames from a short sequence when the head moves to near profile and then back.

\section{GAN Inversion}
\label{sec:gan_inv}

\begin{figure}[t]
  \centering
  \includegraphics[width=\linewidth]{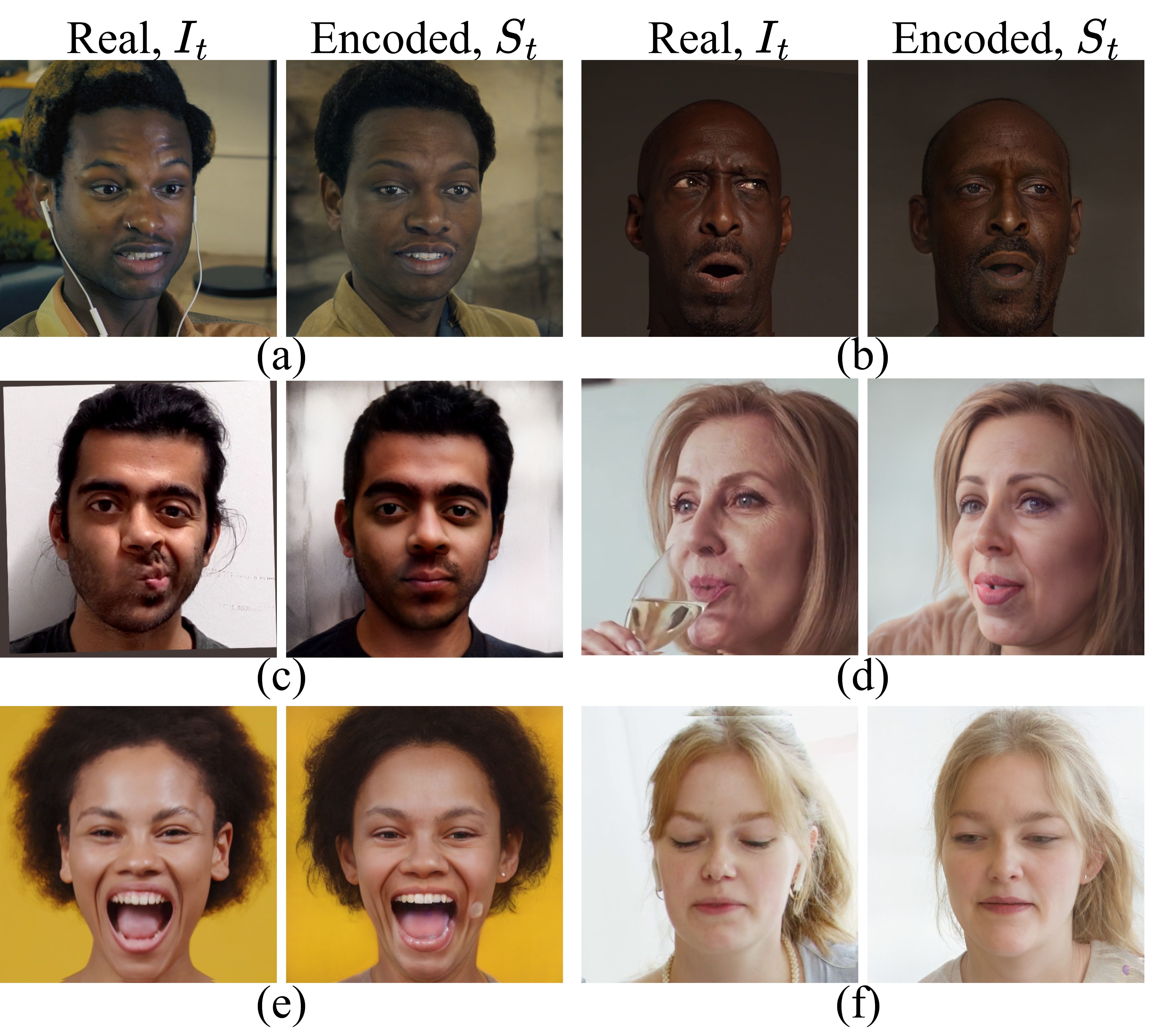}
  \caption{\label{fig:enc_limit} \textbf{Few examples of isolated instances where the e4e encoder fails.} The examples depict: (a) poor identity, (b) incorrect gaze, (c) inability to capture extreme mouth movements, (d) deformations caused due to occlusions, (e) visual artifacts, and (f) flaws in facial features captured (open eyes while closed in real).}
  \vspace{-0.1in}
\end{figure}

Two  factors were considered in choosing an appropriate GAN inversion method: (1) faithful representation of the given image (\ie, minimal reconstruction loss), (2) ability to facilitate latent space edits.  \cite{tov2021designing} suggested that there exists a trade-off between these two factors, \ie, distortion and editability. Generally, inversion is done using a trained encoder and/or an optimization framework. While the former has better editability, it has a comparatively high reconstruction loss and vice versa. Recently,   \cite{roich2021pivotal} proposed to bridge the gap between the two trade-offs by fine-tuning the generator but this adds computational and information transmission costs. 
We chose the e4e encoder, which was designed to facilitate the inversion of real images in proximity to the regions StyleGAN was trained on, thus mitigating the trade-off. 

The e4e encoder \cite{tov2021designing} while producing state-of-the-art results in GAN inversion of real images, has a few failing instances. 
For certain subjects, (\eg, \cref{fig:enc_limit} (a)) the identity of the encoded image deviates considerably from the real frame. In such cases, as we perform the inversion per-frame, there is a tendency for the identity to change across the frames of a single video clip as well. 
The identity change across frames could be due to the poor convergence of the encoder resulting from the existence of a higher per-frame loss due to poor identity.
Additionally, there exist cases where the e4e encoder failed to capture certain facial attributes successfully (\eg, \cref{fig:enc_limit} (b) and (c)) which could be due to the low representation of complex features in the StyleGAN2 training dataset (FFHQ). 
Further, certain visual artifacts and deformations tend to appear in certain cases similar to the examples shown in \cref{fig:enc_limit} (d), (e), and (f), which could be caused due to occlusions (d) and the noisiness in the neighborhood of the inverted $W+$.  

However, the impact of most of these issues on the re-synthesis is mitigated as (1) we anchor our deformations with respect to a single ID-frame that has the highest identity match with the real and (2) utilize a generator fine-tuning stage (PTI\textsubscript{post}) to minimize the identity disparity between the real and synthesized frames.

\section{Identity-Latent Selection}
\label{sec:id_latent}

\begin{figure}[t!]
  \centering
  \includegraphics[width=\linewidth]{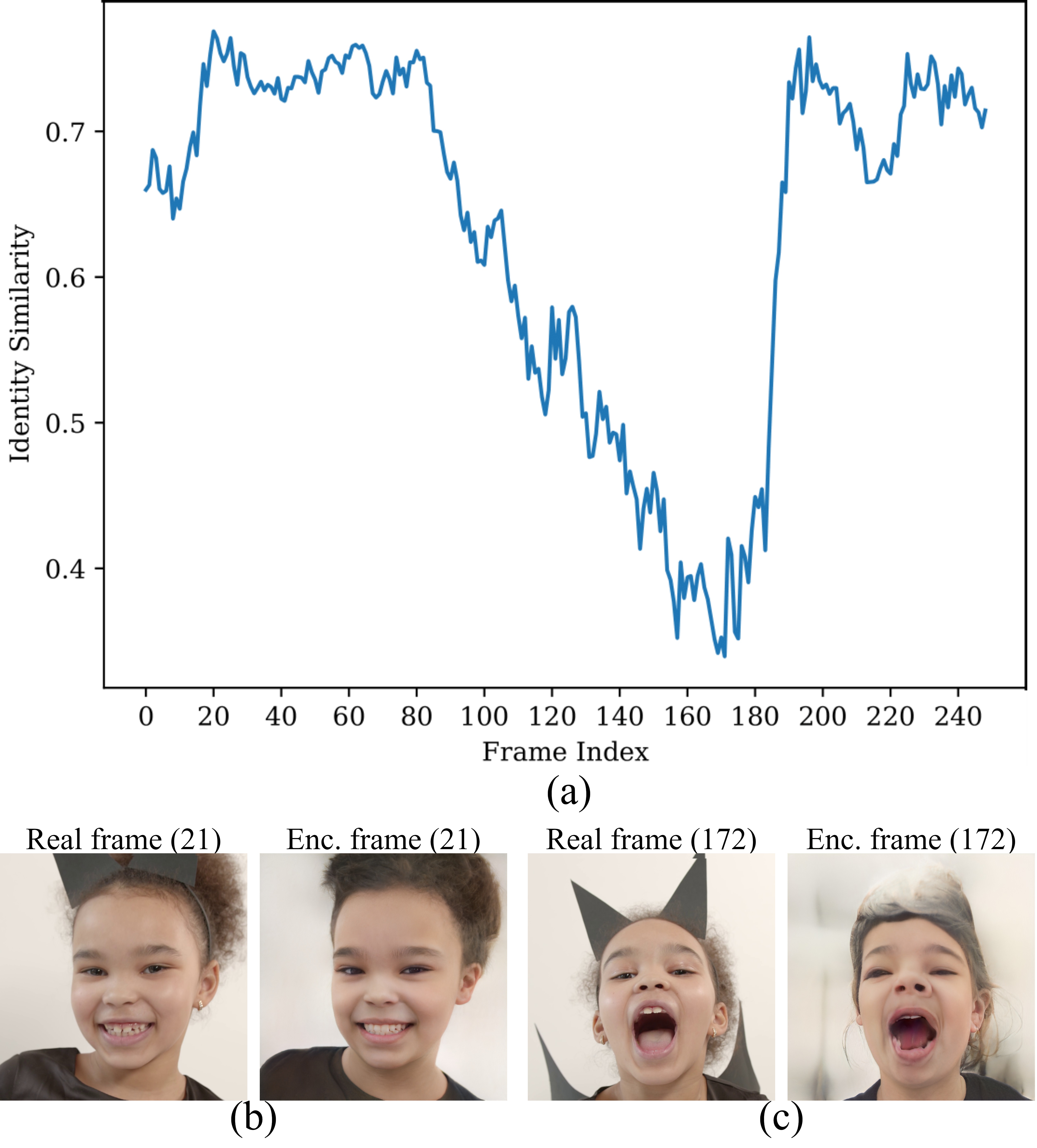} 
  \caption{\label{fig:ID_latent} \textbf{Identifying $L_{ID}$ is based on identity matching using ArcFace.} (a) depicts the identity similarity scores computed between the encoded and real frames. In this case, (b) the best identity is at frame 21 while (c) the worst is at frame 172.}
  \vspace{-0.1in}
\end{figure}

The per-frame inversion creates a series of latents. Depending on the extent of head motion, deformation in StyleGAN2 space is likely. Therefore, the choice of the ID-frame is of great significance as it serves as the base identity for the face and head-pose deformations across the entire sequence of frames.
Hence, we first use ArcFace \cite{deng2019arcface} to compute the similarity between the source and the reconstructed images of the face and then the (1) closest of the face-matches that is also (2) near frontal view of the person, and (3) has no blink is chosen as the representative $L_{ID}$, the basis for re-enactment.
An example plot depicting the variation of the identity similarity (computed based on ArcFace is given in \cref{fig:ID_latent} (a) and the corresponding best and worst ID-frame candidates based on our criteria are shown in \cref{fig:ID_latent} (b).

\section{Facial Attribute Encoding} \label{sec:stylespace}
\subsection{Head-Pose Encoding}

\begin{figure*} [h!]
  \centering
  \includegraphics[width=0.8\linewidth]{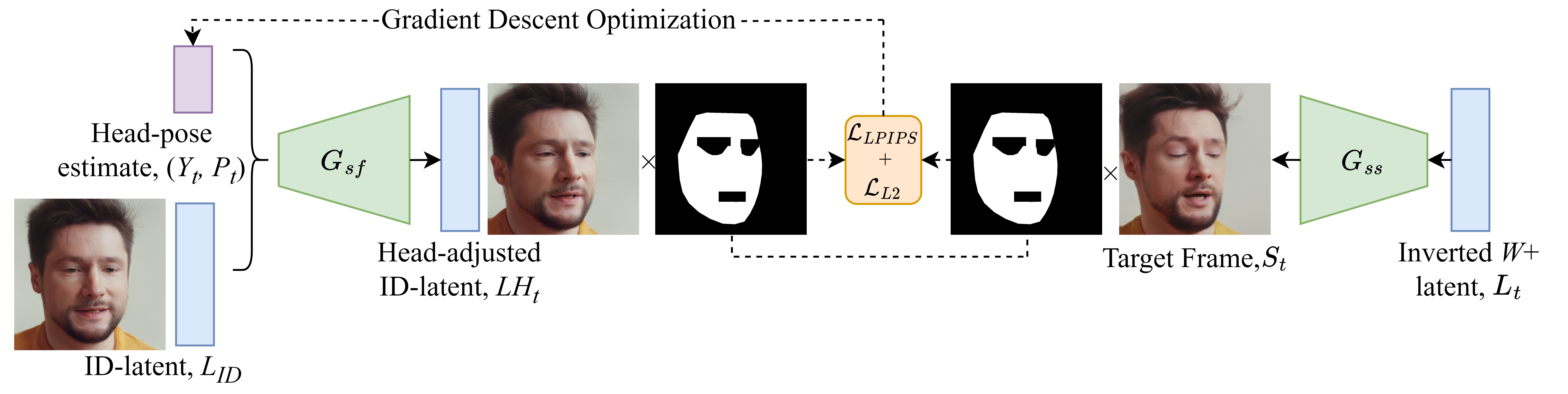}
  \caption{\label{fig:StyleFLow} \textbf{The per-frame head-pose optimization flow.} The head-motion is re-formulated as a head-pose matching problem between a rendered image of the real-frame's encoded latent, $L_t$, and the rendered image of a rotated $L_{ID}$ (defined as $LH_t$) which is solved as a minimization problem employing L2 and  LPIPS losses to search the Yaw-Pitch space using gradient descent. These losses are computed over a masked area of the face excluding the eyes, mouth, and eyebrows since these non-rigid areas are not relevant to 3D head rotations.}
  \vspace{-0.1in}
\end{figure*}

The flow of the head-pose encoding is illustrated in \cref{fig:StyleFLow}. Moreover, to evaluate the significance of our optimization based head-pose encoding approach, we compared our results post head-pose adjustment against the straightforward use of StyleFlow with the $\{Y_t, P_t\}$ parameters computed using \cite{openface2018}. While quantitative results on 6 sample videos were provided in \cref{table:SF_comp}, please refer to the supplementary videos for qualitative comparisons. It could be seen that our approach captures the head-pose well and has a significantly low jitter compared to the straightforward approach with StyleFlow.

\subsection{Choice of StyleSpace Indices}

The significance of the StyleSpace indices selection process as opposed to the algorithm proposed in \cite{wu2021stylespace} is as follows. 
We observed that the StyleSpace, $SS$ representation is not unique. \ie, optimizing 
\begin{equation}
    \min_{\alpha_{inv_t}} \; \mathcal{L}\{ G_{ss}({LH}_t^{ss}+\alpha_t+\alpha_{inv_t}), G_{ss}({LH}_t^{ss})\}
\end{equation}
does not necessarily yield $\alpha_t+\alpha_{inv_t}\approx0$.  
Therefore, as \cite{wu2021stylespace} back propagates to compute the gradient w.r.t. a $SS$ index, the gradients would be less accurate, as the $SS$ indices contributing to an identical facial deformation of two frames would differ (as not unique). Instead, we use a forward approach, perturbing each index separately and computing the corresponding deformation loss, thus directly computing the true gradient (sensitivity in the image space for a unit change of each $SS$ index) which is more accurate. 

\begin{figure*}[t!]
  \centering
  \includegraphics[width=0.81\linewidth]{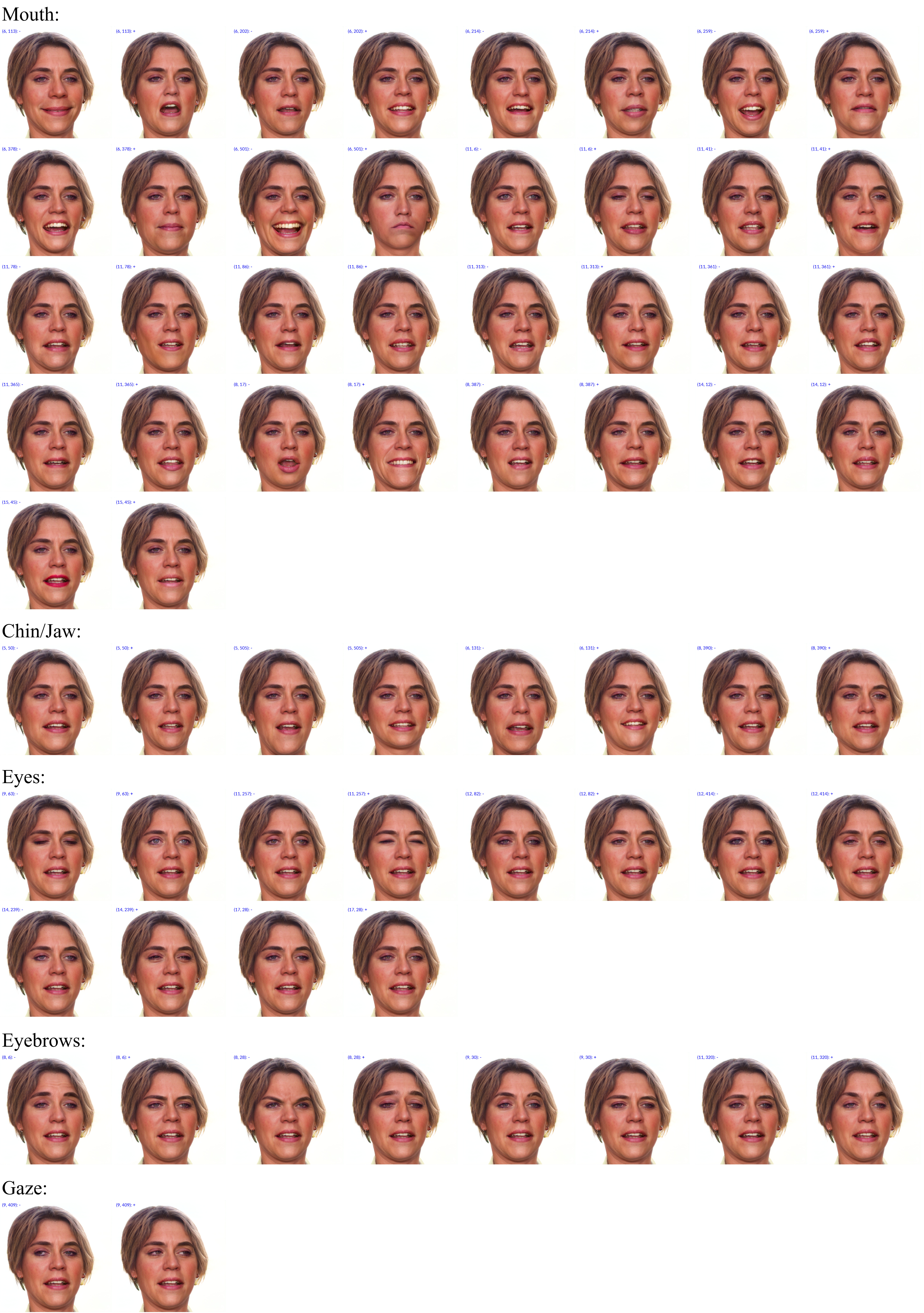}
  \caption{\label{fig:face_deform} \textbf{Example face deformations resulting from manipulation of each StyleSpace index, $(l,c) \in \mathcal{V}$ in the negative (-) and positive (+) directions.} It could be seen that the identity is preserved across all manipulation examples.}
  \vspace{-0.1in}
\end{figure*}

In \cref{fig:face_deform}, we illustrate the facial deformations corresponding to the manipulation of each of the 32 StyleSpace indices tabulated in \cref{tab:SS_idx}. A pair of images marked as $(l,c): +/-$ is included for each StyleSpace index, $(l,c) \in \mathcal{V}$ denoting the sign of the perturbation added to the respective StyleSpace index.  

\begin{table}[t]
\centering
\resizebox{0.95\columnwidth}{!}{
\begin{tabular}{|l|l|}
    \hline
    Facial Attribute, $\mathcal{F}$ & StyleSpace Indices, $\mathcal{V}$ \\
    
    \hline \hline
    
    Mouth &  \vtop{
    \hbox{\strut \{\textbf{6:} 113, 202, 214, 259, 378, 501\},}
    \hbox{\strut \{\textbf{11:} 6, 41, 78, 86, 313, 361, 365\},}
    \hbox{\{\textbf{8:} 17, 387\}, \{\textbf{14:} 12\},\{\textbf{15:} 45\}}
    }\\
    
    \hline
    
    Chin/ Jaw & \vtop{
    \hbox{\{\textbf{5:} 50, 505\}, \{\textbf{6:} 131\}, \{\textbf{8:} 390\}}
    }\\
    
    \hline
    
    Eyes & \vtop{
    \hbox{\{\textbf{9:} 63\}, \{\textbf{11:} 257\}, \{\textbf{12:} 82, 414\},}
    \hbox{\{\textbf{14:} 239\}, \{\textbf{17:} 28\}}
    }\\
    
    \hline
    
    Eyebrows & \vtop{
    \hbox{\{\textbf{8:} 6, 28\}, \{\textbf{9:} 30\}, \{\textbf{11:} 320\}}
    }\\
    
    \hline 
    
    Gaze & \vtop{
    \hbox{\{\textbf{9:} 409\}}
    }\\
    
    \hline
\end{tabular}
}
\caption{\textbf{StyleSpace indices corresponding to the deformation of facial attributes.} The indices take the form of \{$l$: $c_1, c_2, \dots$ \}, where $l$ and $c$ denote respectively the layer index and channel index of the StyleSpace.}
\label{tab:SS_idx}
\end{table}

\begin{figure}[t!]
  \centering
  \includegraphics[width=0.95\linewidth]{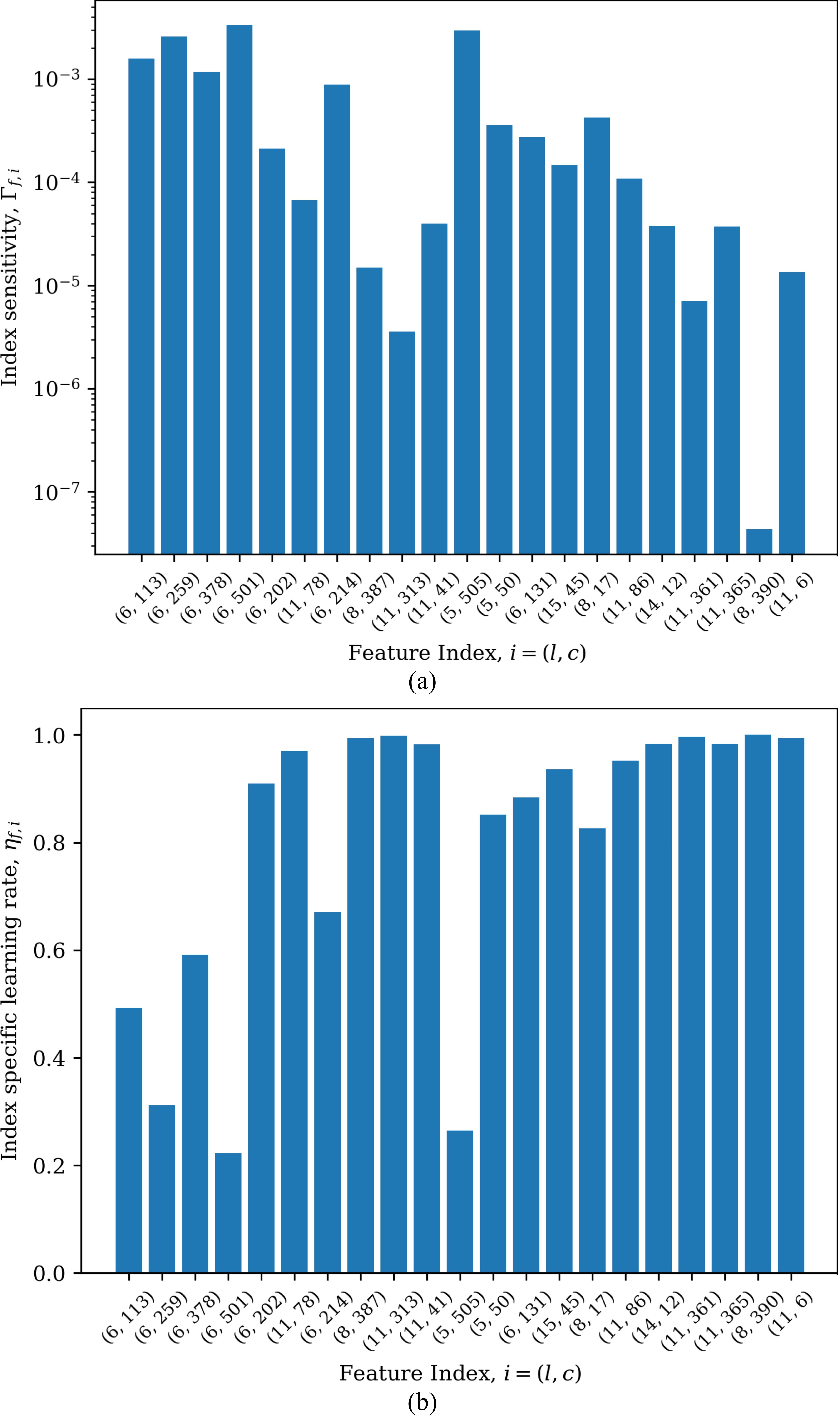}
  \caption{\label{fig:idx_sens} \textbf{(a) Index sensitivity and the corresponding (b) index specific learning rate.} This figure represents values computed for an example subject over the \{mouth+chin/jaw\} indices. }
  \vspace{-0.1in}
\end{figure}

\subsection{Index Specific Learning Rate}
\label{sec:idx_sens}

The variation of index sensitivity computed over the indices corresponding to the \{mouth + chin/jaw\} is shown in \cref{fig:idx_sens} (a). The significant variations seen in the plot make it evident that the index sensitivities cannot be simply ignored and hence, the indices cannot be treated the same during optimization. In order to alleviate the dominance of indices with a higher index sensitivity, we compute an index specific learning rate, $\eta_{f,i}$, $\Gamma_{f,i}$ specified in \cref{eqn:idx_sens} in the main-paper. The $\Gamma_{f,i}$ corresponding to the indices in \cref{fig:idx_sens} (a) are depicted in \cref{fig:idx_sens} (b). It could be seen that the $\eta_{f,i}$ of indices having a higher $\Gamma_{f,i}$ is comparatively lower than the indices of lower $\Gamma_{f,i}$, thus effectively alleviating the dominance. 

\subsection{Details on Optimization}
\label{sec:opt_details}

\RestyleAlgo{ruled}
\begin{algorithm}[t]
\caption{{\label{algo:att_enc}}Optimization Flow for frame t}
\textbf{Inputs}:
\begin{itemize}
    \setlength{\itemsep}{0pt}
    \setlength{\parskip}{0pt}
    \setlength{\parsep}{0pt}
    \item Head-pose adjusted $W+$ latents: $LH_t$ and $LH_{t-1}$
    \item Target frames: $S_{t-1}$ and $S_t$
    \item Rendered frames: $\hat{S}_1$ and $\hat{S}_{t-1}$
    \item StyleSpace of $t-1$: $LH^{ss}_{t-1}$ and $\alpha_{t-1}$
    \item Optimizer $F'$, $N$ number of epochs, and  $G_{ss}$ 
\end{itemize}

\textbf{Initialization:}
\begin{itemize}
    \setlength{\itemsep}{0pt}
    \setlength{\parskip}{0pt}
    \setlength{\parsep}{0pt}
    \item Obtain the StyleSpace latent, $LH^{ss}_t = \mathcal{A}(LH_t)$
    \item Initialize $LH^{ss}_t(l,c),\, \forall i = (l,c) \in \mathcal{V}$ 
    \item Initialize $\alpha_t = [0, \dots, 0]$
    \item Compute the index-specific learning rates, $\eta_{f,i}$ \\
    \qquad $\eta = \{\eta_{f,i}; \, \forall f \in \mathcal{F}, i \in \mathcal{V}\}$
\end{itemize}
\textbf{Optimization:}

\For{n = [1:N]}{
$\hat{S}_{t} = G_{ss}\{LH^{ss}_t + \alpha_{t} \mathbbm{1}_i\}$\\  \quad where $\mathbbm{1}_i = \{1$ when $(l,c) \in \mathcal{V};$ $0$ elsewhere\}\\
$\mathcal{L} = \mathcal{L}\{\hat{S}_1,\,\hat{S}_{t-1}, \hat{S}_t,\,S_{t-1},\,S_{t}\}$ \\
$\alpha_t \gets \alpha_t - \eta F'(\nabla_{\alpha_t}\mathcal{L}, \alpha_t)$
}

\textbf{Output:}
\begin{itemize}
    \item 32-dimensional $\alpha_t$
\end{itemize}
\end{algorithm}

The face deformation attribute encoding algorithm in \cref{sub:fac_attr} of the main paper is presented in \cref{algo:att_enc}. The AdamW \cite{loshchilov2018decoupled} optimizer with AMSGrad \cite{reddi2018convergence} was utilized with an initial learning rate of $\eta = \{\eta_{f,i}; \, \forall f \in \mathcal{F}, i \in \mathcal{V}\}$, $(\beta_1,\, \beta_2) = (0.9,\,0.999)$, and $\epsilon=1e^{-8}$. The optimization was over 100 epochs ($N = 100$) and the learning rate was decayed every $10$ epochs with a decaying factor of $0.8$ using a learning rate scheduler for improved convergence. The optimization was approximately 1 min./frame on a single GTX1080Ti GPU. Additional details on the loss terms defined in \cref{eq:total_loss,eq:loss_m,eq:loss_e,eq:loss_p} of the main-paper are given below. 

$\boldsymbol{\mathcal{L}_{LPIPS}}$: The LPIPS loss \cite{zhang2018perceptual}, which is known to learn perceptual similarities well \cite{guan2020collaborative,richardson2021encoding}, was used to capture the structural details of the facial attributes between $S_t$ and $\hat{S}_t$. Nevertheless, $\mathcal{L}_{LPIPS}$ was not used in solving for the gaze ($\mathcal{L}_p$) as it is invariant to slight spatial changes and hence introduces a slight jitter when used. 

$\boldsymbol{\mathcal{L}_{L2}}$: This denotes the L2 norm between the $S_t$ and $\hat{S}_t$, and enables precise reconstruction (\eg, the case of gaze). 

$\boldsymbol{\mathcal{L}_{ID}}$: To mitigate the risk of changing the identity of the subject across frames while optimizing over the latent space, the identity loss is in place as a regularization term. This is computed between $\hat{S}_1$ and $\hat{S}_t$.  

$\boldsymbol{\mathcal{L}_{FP}}$: As we optimize over 32 indices in parallel, we noted  occasional nose, mouth, and chin/jaw deformations. To discourage unwarranted deformations, the Face-Parsing loss, which is the L2 norm of the difference between the masked face-parsing scores \cite{yu2018bisenet} 
of the rendered and target frames, is used instead of facial-landmark coordinates loss (\eg, \cite{openface2018}). Face-parsing scores  facilitate the gradient flow through the optimization and are more precise and stable across the frames. 
\begin{align}
    \mathcal{L}_{FP} = ||FP(\hat{S}_t) * M - FP(S_t) * M||_2
\end{align}
where function $FP(\cdot)$ yields face-parsing scores and $M$ denotes the binary mask of the face. 

$\boldsymbol{\mathcal{L}_{IF}}$: The inter-frame loss is a derivation of the Frame Difference-Based (FDB) loss proposed  in \cite{xu2021frame}, to enforce  temporal coherence between frames. We minimize this loss along with the other spatial losses to avoid  enforcing temporal continuity posteriori. Provided the target video is temporally coherent, this loss is based on the concept that the image space and feature space differences between consecutive frames embed  the temporal coherence. We use LPIPS and L2 losses to compute differences in the feature and image spaces, respectively. 
\begin{align}
    \mathcal{L}_{IF} &= \mathcal{L}_{IF\_LPIPS} + \mathcal{L}_{IF\_L2} \\
    \mathcal{L}_{IF\_*} &= \mathcal{L}_{*}\{S_t, S_{t-1}\} -  \mathcal{L}_{*}\{\hat{S}_t, \hat{S}_{t-1}\}
\end{align}
where $*$ denotes either LPIPS or L2.

\section{Experiments and Results}
\label{sec:results}

\subsection{Dataset}\label{sec:dataset}
As stated in \cref{sub:dataset_eval} of the main-paper, we compose a dataset consisting of video clips of 4K resolution sourced from the site \url{www.pexels.com}. The videos were chosen such that diverse subjects belonging to various ethnicities, age groups, and having different facial geometries, performing significant head-pose movements and facial deformations (both expressions and speech) were included. The results were computed based on 150 videos chosen from the dataset, with a mean of 304 frames, a minimum of 100 frames, and a maximum of 1000 frames. 

\subsection{Evaluation Metrics}\label{sec:metrics}
\begin{figure*}[h!]
  \centering
  \includegraphics[width=\linewidth]{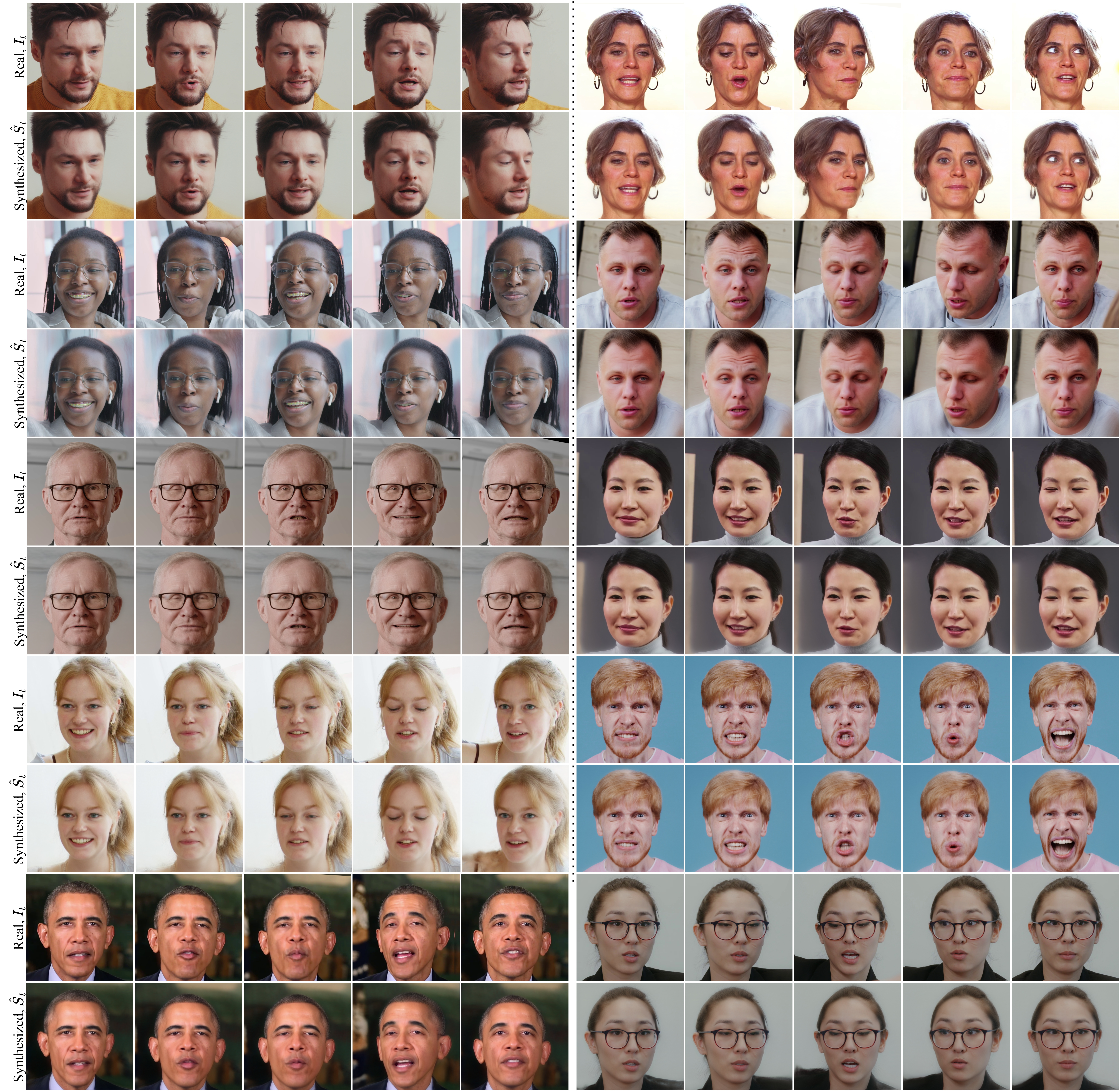}
  \caption{\label{fig:supp_resyn} \textbf{Additional qualitative examples of re-synthesis} demonstrating the versatility of our algorithm.  
  }
\end{figure*}
The following metrics were used for the quantitative evaluation of our re-enactment videos in comparison with baselines, which are tabulated in \cref{tab:re-synthesis,tab:puppeteering} of the main-paper.
 
\textbf{Mean L1-distance, L1}: The per-pixel L1-distance was averaged across pixels, channels, and frames to obtain the score. The pixel values of the input images were in the range of [0,255].

\textbf{Learned Perceptual Image Patch Similarity Loss, LPIPS}: The metric was computed per-frame using the original implementation of \cite{zhang2018perceptual} computed using the feature space of AlexNet \cite{krizhevsky2012imagenet}. 

\textbf{Identity Loss, $\mathcal{L}_{ID}$}: The identity loss was computed using,
\begin{equation}
    \mathcal{L}_{ID} = 1 - \langle \phi(S_t) , \phi(\hat{S}_t) \rangle
\end{equation}
where $\phi$ represents the pretrained ArcFace network and $\langle \cdot , \cdot \rangle$ denotes the cosine similarity. While in re-synthesis (\cref{tab:re-synthesis} in the main-paper) the loss was computed between the synthesized frame and the real frame, for puppeteering (\cref{tab:puppeteering} in the main-paper) the loss was computed between each frame and the puppet's ID-frame. 

\begin{figure}[t]
  \centering
\includegraphics[width=\linewidth]{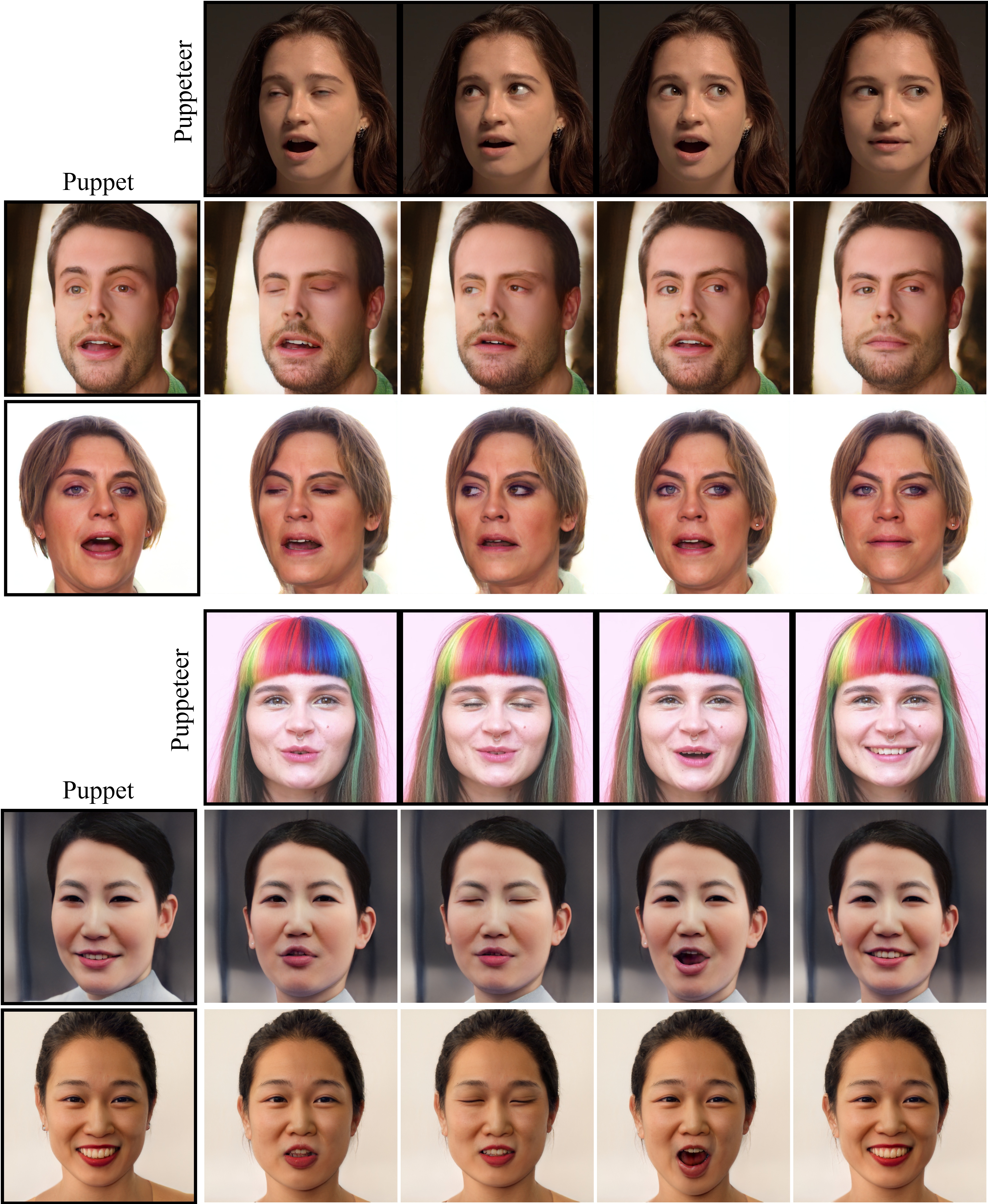}
  \caption{\label{fig:supp_p2p} \textbf{Additional qualitative examples of puppeteering} generated by applying the encoded parameters computed for the puppeteer through our encoding algorithm onto the ID-latent of the puppet.}
\end{figure}

\textbf{Peak Signal to Noise Ratio, PSNR}: This was computed using the built-in function of python's scikit-image package using images having pixel values in the range [0,255].

\textbf{Fr\'{e}chet Inception Distance, FID}: This metric, which is used to measure the photo-realism between two datasets, was computed based on the original implementation of \cite{heusel2017gans} with a batch size of 100. Note: The input images are rescaled to $299\times299$ at the input of the inception network. 

\textbf{Fr\'{e}chet Video Distance, FVD}: The spatio-temporal perceptual score measured through FVD was computed using the original implementation of \cite{unterthiner2018towards}. Video fragments of length 120 frames were scored with a batch size of 8 and averaged to obtain the final FVD score due to resource limitations. Note: The frames are rescaled to $224\times224$ by the algorithm. 

\textbf{Fr\'{e}chet Video Distance - Mouth, FVD$_M$}: Similar to FVD, with the exception of the metric being scored over the masked area of the mouth region.

\textbf{Action Unit, Gaze, Pose Correlations, $\mathbf{\rho_{\scaleto{AU}{3pt}}}$, $\mathbf{\rho_{\scaleto{GZ}{3pt}}}$, $\mathbf{\rho_{\scaleto{pose}{3pt}}}$}: These metrics measure the time-series correlation between the Action Unit activations, Gaze angles, and Yaw and Pitch angles respectively, which are computed using OpenFace 2.0 \cite{openface2018} of the synthesized and the reference sequences. 
These provide an insight into how well the facial deformations ($\mathbf{\rho_{\scaleto{AU}{3pt}}}$), eye motion ($\mathbf{\rho_{\scaleto{GZ}{3pt}}}$), and pose ($\mathbf{\rho_{\scaleto{pose}{3pt}}}$) are captured by the algorithm in a spatio-temporal sense.

Note: All metrics except FVD, were computed per frame and averaged across all the frames. Further, except for identity loss and correlation metrics, all other metrics were computed over a masked-out region of the reference face of each frame. 

\subsection{Video Results}

The additional examples of video re-synthesis and puppeteering depicted in \cref{fig:supp_resyn} and \cref{fig:supp_p2p} respectively reaffirm the versatility of our approach. Video examples comparing the state-of-the-art approaches could be viewed on the project page. In comparison to our results, visual artifacts, lack of sharpness, and incorrect pose and facial deformations could be observed in the re-synthesis and puppeteering examples of the baseline approaches. 

\subsection{Limitations}

There are multiple scenarios where latent-based video encoding may fail: 
(1) due to limitations inherited from StyleGAN2 (\eg, fixed resolution, entanglements, alignment requirements, texture sticking), 
(2) during pre-processing if the face is misaligned with respect to StyleGAN2 expectations,
(3) extreme facial deformations and profile views,  stemming from the low representation in the FFHQ dataset used in training StyleGAN2,
(4) possible identity drift in editing  StyleFlow or StyleSpace,
(5) wearables such as eyeglasses can be challenging in some cases due to remaining latent space entanglement,
(6) both latent-space inversion and editing are sensitive to occlusions.

\section{Potential Negative Societal Impact}\label{sec:negative}
Since the proposed pipeline successful captures the fine, detailed, and expressive facial attributes, it improves the realism of face re-enactment. Thus, our model could be misused to create re-enactments with ill-intent (\eg, defamation) and we strongly oppose such malicious use. The research on detection of DeepFakes have progressively advanced as well \cite{wang2022gan,juefei2022countering,zhang2022exposing,gangan2022distinguishing}, and the data from our model could be used to improve such methods, thus reducing the potential negative societal impact.   

\end{document}